\documentclass[11pt,a4paper]{article}
\usepackage[hyperref]{acl2021}
\usepackage{times}
\usepackage{latexsym}
\usepackage{graphicx}
\usepackage{subcaption}
\usepackage{booktabs}
\usepackage{soul}
\usepackage{tabularx}

\usepackage{microtype}

\aclfinalcopy 

\newcommand{\ignore}[1]{}

\title{On the Efficacy of Adversarial Data Collection for Question Answering: \\ 
Results from a Large-Scale Randomized Study
}

\author{
Divyansh Kaushik$^\dagger$, Douwe Kiela$^\ddagger$, Zachary C. Lipton$^\dagger$, Wen-tau Yih$^\ddagger$\\[1ex]
$^\dagger$ Carnegie Mellon University; $^\ddagger$ Facebook AI Research\\
  \texttt{\{dkaushik,zlipton\}@cmu.edu}, \hspace{2mm} \texttt{\{dkiela,scottyih\}@fb.com}\\}

\date{}

\begin{document}
\maketitle
\begin{abstract}
In \emph{adversarial data collection} (ADC),
a human workforce interacts 
with a model in real time,
attempting to produce examples 
that elicit incorrect predictions. 
Researchers hope that models trained 
on these more challenging datasets
will rely less on superficial patterns,
and thus be less brittle. 
However, despite ADC's intuitive appeal,
it remains unclear 
when training on adversarial datasets 
produces more robust models.
In this paper, we conduct
a large-scale controlled study 
focused on question answering, 
assigning workers at random 
to compose questions either
(i) adversarially (with a model in the loop);
or (ii) in the standard fashion (without a model).
Across a variety of models and datasets, 
we find that models trained on adversarial data 
usually perform better 
on other adversarial datasets
but worse on a diverse collection of 
out-of-domain evaluation sets. 
Finally, we provide a qualitative analysis
of adversarial (vs standard) data,
identifying key differences 
and offering guidance for future research.\footnote{Data collected during this study is publicly available at \href{https://github.com/facebookresearch/aqa-study}{https://github.com/facebookresearch/aqa-study}.}

\end{abstract}

\section{Introduction}
\begin{figure*}[!t]
\centering
    \includegraphics[width=0.801\textwidth]{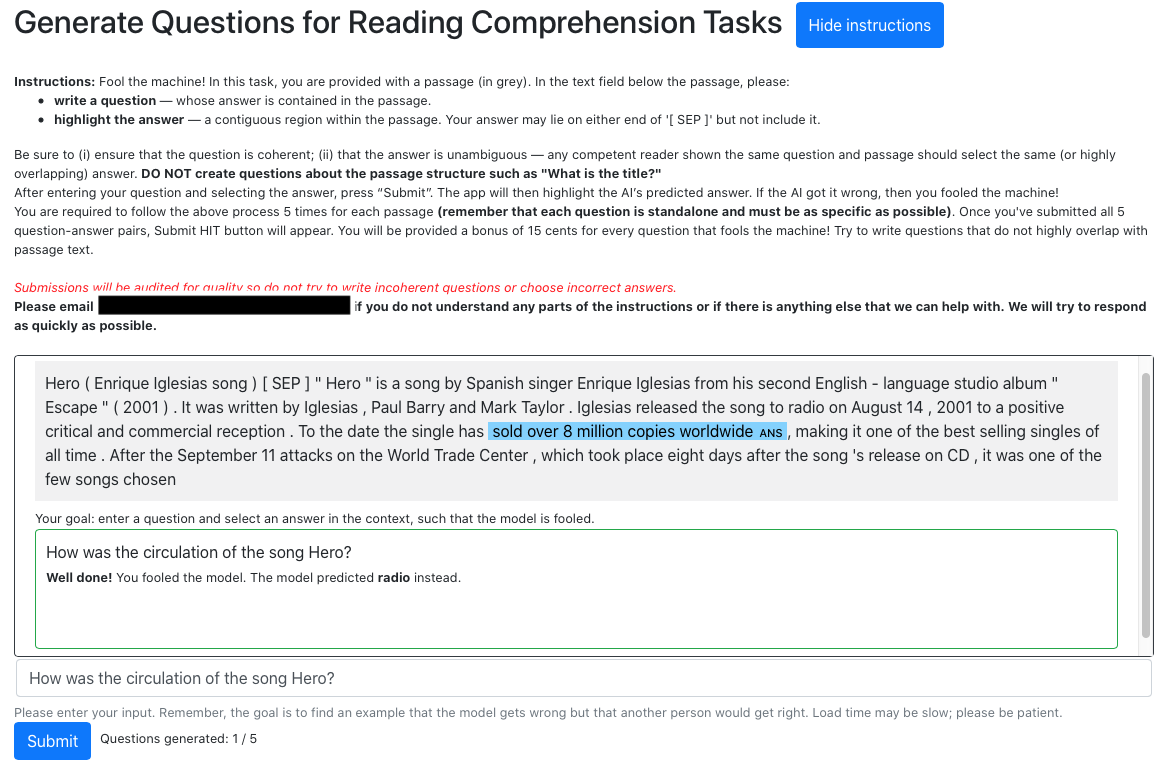}
\caption{
Platform shown to workers generating questions in the ADC setting.
\label{fig:annotation_platform2}}
\end{figure*}
\label{sec:introduction}
Across such diverse natural language processing (NLP) tasks 
as natural language inference 
\citep[NLI;][]{poliak2018hypothesis, gururangan2018annotation},
question answering \citep[QA;][]{kaushik2018much},
and sentiment analysis \citep{Kaushik2020Learning},
researchers have discovered 
that models can succeed on popular benchmarks 
by exploiting spurious associations 
that characterize a particular dataset
but do not hold more widely.
Despite performing well 
on independent and identically distributed (i.i.d.) data,
these models are liable 
under plausible domain shifts.
With the goal of providing 
more challenging benchmarks
that require this stronger 
form of generalization,
an emerging line of research 
has investigated \emph{adversarial data collection} (ADC),
a scheme in which a worker 
interacts with a model (in real time),
attempting to produce examples
that elicit incorrect predictions 
\citep[e.g.,][]{dua2019drop,nie2020adversarial}.
The hope is that by identifying 
parts of the input domain 
where the model fails 
one might make the model more robust.
Researchers have shown that
models trained on ADC perform better 
on such adversarially collected data
and that with successive rounds of ADC,
crowdworkers are less able to fool the models
\citep{dinan2019build}.

While adversarial data may indeed provide
more challenging benchmarks,
the process and its actual benefits 
vis-a-vis tasks of interest 
remain poorly understood,
raising several key questions:
(i) do the resulting models typically generalize better
out of distribution compared to standard data collection (SDC)?;
(ii) how much can differences between ADC and SDC 
be attributed to the way workers behave 
when attempting to fool models,
regardless of whether they are successful?
and (iii) what is the impact of training models
on adversarial data only,
versus using it as a data augmentation strategy?

In this paper, we conduct a large-scale
randomized controlled study 
to address these questions.
Focusing our study on span-based question answering
and a variant of the Natural Questions dataset~\citep[NQ;][]{lee2019latent,karpukhin2020dense},
we work with two popular
pretrained transformer architectures---BERT\textsubscript{large} 
\citep{devlin2019bert} and 
ELECTRA\textsubscript{large} \citep{Clark2020ELECTRA}---each 
fine-tuned on $23.1k$ examples.
To eliminate confounding factors 
when assessing the impact of ADC,
we randomly assign the crowdworkers
tasked with generating questions 
to one of three groups: 
(i) with an incentive to fool the BERT model; 
(ii) with an incentive to fool the ELECTRA model; 
and (iii) a standard, non-adversarial setting
(no model in the loop).
The pool of contexts is the same for each group
and each worker is asked to generate 
five questions for each context that they see. 
Workers are shown similar instructions 
(with minimal changes), 
and paid the same base amount.

We fine-tune three models
(BERT, RoBERTa, and ELECTRA) 
on resulting datasets 
and evaluate them on held-out test sets,
adversarial test sets from prior work \citep{bartolo2020beat}, 
and $12$ MRQA \citep{fisch2019mrqa} datasets.
For all models, we find that 
while fine-tuning on adversarial data 
usually leads to better performance 
on (previously collected) adversarial data,
it typically leads 
to worse performance
on a large, diverse collection  
of out-of-domain datasets 
(compared to fine-tuning on standard data).
We observe a similar pattern
when augmenting the existing dataset
with the adversarial data.
Results on an extensive collection
of out-of-domain evaluation sets
suggest that ADC training data 
does not offer clear benefits 
vis-\`{a}-vis robustness under distribution shift.

To study the differences
between adversarial and standard data, 
we perform a qualitative analysis,
categorizing questions based on 
a taxonomy \citep{hovy2000question}.
We notice that more questions in the ADC dataset
require numerical reasoning 
compared to the SDC sample.
These qualitative insights may offer 
additional guidance to future researchers.

\section{Related Work}
\label{sec:related}
In an early example of model-in-the-loop
data collection,
\citet{zweig2012challenge}
use $n$-gram language models
to suggest candidate incorrect answers 
for a fill-in-the-blank task.
\citet{richardson2013mctest} 
suggested ADC for QA 
as proposed future work, 
speculating that it might
challenge state-of-the-art models.
In the \emph{Build It Break It, The Language Edition} 
shared task \citep{ettinger2017towards}, 
teams worked as \emph{builders} (training models) 
and \emph{breakers} 
(creating challenging examples for subsequent training)
for sentiment analysis and QA-SRL.

Research on ADC has picked up recently,
with \citet{chen2019codah} tasking crowdworkers 
to construct multiple-choice questions 
to fool a BERT model
and \citet{wallace2019trick} 
employing Quizbowl community members
to write Jeopardy-style questions 
to compete against QA models.
\citet{zhang2018record} automatically generated 
questions from news articles,
keeping only those questions 
that were incorrectly answered by a QA model. 
\citet{dua2019drop} and \citet{dasigi2019quoref}
required crowdworkers to submit only questions 
that QA models answered incorrectly.
To construct FEVER $2.0$ \citep{thorne2019fever2},
crowdworkers were required to fool
a fact-verification system 
trained on the FEVER \citep{thorne2018fever} dataset.
Some works explore ADC over multiple rounds,
with adversarial data from one round
used to train models 
in the subsequent round.
\citet{yang2018mastering} ask workers
to generate challenging datasets
working first as adversaries and later as collaborators.
\citet{dinan2019build} build on their work,
employing ADC to address offensive language identification.
They find that over successive rounds of training,
models trained on ADC data 
are harder for humans to fool 
than those trained on standard data.
\citet{nie2020adversarial} applied ADC 
for an NLI task over three rounds,
finding that training for more rounds
improves model performance on adversarial data,
and observing improvements 
on the original evaluations set
when training on a mixture 
of original and adversarial training data.
\citet{williams2020anlizing} conducted an
error analysis of model predictions on the datasets 
collected by \citet{nie2020adversarial}.
\citet{bartolo2020beat} studied 
the empirical efficacy of ADC 
for SQuAD \citep{rajpurkar2016squad},
observing improved performance 
on adversarial test sets
but noting that trends vary 
depending on the models used 
to collect data and to train.
Previously, \citet{lowell2019practical} 
observed similar issues in active learning, 
when the models used to acquire data 
and for subsequent training differ.
\citet{yang2018hotpotqa, zellers2018swag, zellers2019hellaswag} 
first collect datasets
and then filter examples 
based on predictions from a model. 
\citet{paperno2016lambada} apply a similar procedure
to generate a language modeling dataset (LAMBADA). 
\citet{Kaushik2020Learning, kaushik2021explaining}
collect counterfactually augmented data (CAD)
by asking crowdworkers to edit existing documents 
to make counterfactual labels applicable,
showing that models trained on CAD
generalize better out-of-domain.

Absent further assumptions,
learning classifiers 
robust to distribution shift
is impossible \citep{david2010impossibility}. 
While few NLP papers on the matter
make their assumptions explicit,
they typically proceed under the implicit assumptions
that the labeling function is deterministic 
(there is one right answer),
and that \emph{covariate shift} 
\citep{shimodaira2000improving} applies
(the labeling function $p(y|x)$ 
is invariant across domains). 
Note that neither condition is 
generally true of prediction problems.
For example, faced with label shift 
\citep{scholkopf2012causal, lipton2018detecting}
$p(y|x)$ can change across distributions,
requiring one to adapt the predictor to each environment.

\section{Study Design}
In our study of ADC for QA,
each crowdworker is shown a short passage 
and asked to create~$5$ questions 
and highlight answers 
(spans in the passage, 
see Fig.~\ref{fig:annotation_platform2}).
We provide all workers with the same base pay
and for those assigned to ADC, 
pay out an additional bonus 
for each question that fools the QA model.
Finally, we field a different set of workers
to validate the generated examples.

\paragraph{Context passages}
\label{sec:passages}
For context passages, 
we use the first $100$ words 
of Wikipedia articles.
Truncating the articles keeps 
the task of generating questions 
from growing unwieldy.
These segments typically contain an overview,
providing ample material for factoid questions. 
We restrict the pool of candidate contexts by leveraging a variant of the Natural Questions dataset~\cite{kwiatkowski2019natural,lee2019latent}. 
We first keep only a subset of $23.1k$ question/answer pairs for which the context passages are the first $100$ words of Wikipedia articles\footnote{We 
used the data prepared by~\citet{karpukhin2020dense},
available at \href{https://www.github.com/facebookresearch/DPR}{https://www.github.com/facebookresearch/DPR}.}.
From these passages, we sample $10k$ at random for our study.

\paragraph{Models in the loop} 
We use BERT\textsubscript{large} \citep{devlin2019bert} 
and ELECTRA\textsubscript{large}~\citep{Clark2020ELECTRA} models 
as our adversarial models in the loop, 
using the implementations provided 
by \citet{wolf2020transformers}.
We fine-tune these models 
for span-based question-answering,
using the $23.1k$ training examples
(subsampled previously)
for $20$ epochs,
with early-stopping based on word-overlap 
F1\footnote{Word-overlap F1 and Exact Match (EM) metrics 
introduced in \citet{rajpurkar2016squad}
are commonly used to evaluate performance 
of passage-based QA systems,
where the correct answer 
is a span in the given passage.}
over the validation set.
Our BERT model achieves 
an EM score of $73.1$ 
and an F1 score of $80.5$ 
on an i.i.d.~validation set. 
The ELECTRA model performs 
slightly better, obtaining an 
$74.2$ EM and $81.2$ F1 on the same set.

\paragraph{Crowdsourcing protocol}
We build our crowdsourcing platform
on the Dynabench interface~\citep{kiela2021dynabench}
and use Amazon's Mechanical Turk 
to recruit workers to write questions.
To ensure high quality,
we restricted the pool to U.S. residents
who had already completed at least $1000$ HITs 
and had over $98\%$ HIT approval rate.
For each task, we conducted 
several pilot studies
to gather feedback from crowdworkers
on the task and interface.
We identified median time taken by workers 
to complete the task 
in our pilot studies
and used that to design
the incentive
structure for the main task.
We also conducted multiple studies 
with different variants of instructions 
to observe trends in the quality of questions
and refined our instructions 
based on feedback from crowdworkers.
Feedback from the pilots also guided
improvements to our crowdsourcing interface. 
In total, $984$ workers took part in the study, 
with $741$ creating questions.
In our final study,
we randomly assigned workers 
to generate questions in the following ways: 
(i) to fool the BERT baseline; 
(ii) to fool the ELECTRA baseline;
or
(iii) without a model in the loop.
Before beginning the task, 
each worker completes an onboarding process 
to familiarize them with the platform.
We present the same set of passages to workers
regardless of which group they are assigned to,
tasking them with generating 
$5$ questions for each passage.

\paragraph{Incentive structure}
During our pilot studies, we found 
that workers spend $\approx 2$--$3$ minutes 
to generate $5$ questions.
We provide workers with the same base 
pay---$\$0.75$ per HIT---(to
ensure compensation at a $\$15$/hour rate).
For tasks involving a model in the loop, 
we define a model prediction to be \emph{incorrect} 
if its F1 score is less than $40\%$,
following the threshold set by \citet{bartolo2020beat}.
Workers tasked with fooling the model
receive bonus pay of $\$0.15$ 
for every question that leads 
to an incorrect model prediction.
This way, a worker can double their pay
if all $5$ of their generated questions
induce incorrect model predictions.

\paragraph{Quality control}
Upon completion of each batch 
of our data collection process,
we presented~\mbox{$\approx 20\%$}
of the collected questions 
to a fourth group of crowdworkers
who were tasked with validating whether 
the questions were answerable 
and the answers were correctly labeled.
In addition, we manually verified 
a small fraction of 
the collected question-answer pairs.
If validations of at least~$20\%$ of the examples
generated by a particular worker were incorrect,
their work was discarded in its entirety.
The entire process, 
including the pilot studies cost~$\approx \$50k$ and
spanned a period of seven months.
Through this process, we collected 
over~$150k$ question-answer pairs 
corresponding to the~$10k$ contexts 
($50k$ from each group)
but the final datasets are much smaller,
as we explain below.

\begingroup
\setlength{\tabcolsep}{3.25pt}
\begin{table}[!t]
    \centering
    \small
    \begin{tabular}{c c c c c c c}
        \toprule
      Resource  & \multicolumn{3}{c}{Num. Passages} & \multicolumn{3}{c}{Num. QA Pairs} \\
      & Train & Val & Test & Train & Val & Test \\
      \midrule
        BERT & 3,412 & 992 & 1,056 & 11,330 & 1,130 & 1,130 \\
        ELECTRA & 3,925 & 1,352 & 1,352 & 14,556 & 1,456 & 1,456 \\
        \bottomrule
    \end{tabular}
    \caption{Number of unique passages and question-answer pairs for each data resource.}
    \label{tab:data_stats}\vspace{-5mm}
\end{table}
\endgroup

\begingroup
\begin{table*}[t]
  \centering
  \tiny
  \begin{tabular}{ l c c c c c c c c}
    \toprule
    Evaluation set $\rightarrow$ & \multicolumn{2}{c}{BERT\textsubscript{fooled}} & \multicolumn{2}{c}{BERT\textsubscript{random}} & \multicolumn{2}{c}{SDC} & \multicolumn{2}{c}{Original Dev.} \\
    Training set $\downarrow$ & EM & F1 & EM & F1 & EM & F1 & EM & F1\\
     \midrule
        \multicolumn{9}{c}{Finetuned model: BERT\textsubscript{large}}\\
    \midrule
    Original (O; 23.1k) & $0.0$ & $17.1$ & $29.6$ & $45.2$ & $32.5$ & $49.1$ & $73.3$ & $80.5$\\
    Original (11.3k) & $8.4_{0.9}$ & $18.7_{0.6}$ & $28.8_{0.5}$ & $42.7_{0.9}$ & $33.1_{0.7}$ & $48.6_{1.1}$ & $66.1_{0.3}$ & $74.2_{0.4}$ \\
    \midrule
    BERT\textsubscript{fooled} (F; 11.3k) & $34.4_{5.1}$ & $57.0_{5.7}$ & $44.0_{8.8}$ & $61.7_{8.2}$ & $47.5_{10.0}$ & $66.8_{8.6}$ & $34.5_{2.6}$ & $47.9_{3.3}$\\
    BERT\textsubscript{random} (R; 11.3k) & $37.7_{2.7}$ & $58.9_{2.5}$ & $57.0_{4.5}$ & $73.9_{3.5}$ & $62.4_{4.5}$ & $79.7_{3.1}$ & $46.4_{3.1}$ & $60.6_{3.8}$\\
    SDC (11.3k) & $33.6_{0.3}$ & $54.4_{0.4}$ & $57.6_{0.6}$ & $74.5_{0.4}$ & $\mathbf{68.6_{0.5}}$ & $\mathbf{84.2_{0.3}}$ & $48.6_{1.6}$ & $62.3_{1.9}$\\
    \midrule
    O + F (34.4k) & $\mathbf{39.9_{0.8}}$ & $\mathbf{61.7_{0.5}}$ & $50.6_{0.9}$ & $68.5_{0.9}$ & $52.6_{1.4}$ & $71.8_{1.1}$ & $72.2_{0.4}$ & $79.8_{0.6}$\\
    O + R  (34.4k) & $38.1_{0.5}$ & $58.8_{0.6}$ & $57.9_{1.0}$ & $74.8_{0.5}$ & $62.6_{0.5}$ & $80.2_{0.3}$ & $72.5_{0.5}$ & $80.2_{0.3}$\\
    O + SDC  (34.4k) & $33.4_{0.4}$ & $54.5_{0.6}$ & $60.6_{4.4}$ & $77.2_{3.6}$ & $\mathbf{69.0_{0.3}}$ & $\mathbf{84.3_{0.3}}$ & $72.1_{0.2}$ & $79.8_{0.2}$\\
    \midrule
    \multicolumn{9}{c}{Finetuned model: RoBERTa\textsubscript{large}}\\
    \midrule
    Original (O; 23.1k) & $7.3$ & $16.7$ & $28.6$ & $44.5$ & $32.7$ & $50.1$ & $73.5$ & $80.5$\\
    Original (11.3k) & $4.5_{0.4}$ & $10.8_{1.1}$ & $17.5_{0.9}$ & $26.7_{2.0}$ & $19.5_{2.1}$ & $30.0_{3.2}$ & $70.6_{0.3}$ & $78.5_{0.4}$ \\
    \midrule
    BERT\textsubscript{fooled} (F; 11.3k) & $\mathbf{49.2_{0.5}}$ & $\mathbf{71.2_{0.7}}$ & $64.9_{1.3}$ & $81.3_{1.1}$ & $67.9_{1.5}$ & $84.8_{1.0}$ & $41.4_{1.0}$ & $55.1_{1.1}$\\
    BERT\textsubscript{random} (R; 11.3k) & $48.0_{0.4}$ & $69.8_{0.4}$ & $\mathbf{70.3_{0.7}}$ & $\mathbf{85.3_{0.4}}$ & $72.5_{0.4}$ & $87.8_{0.1}$ & $50.6_{0.8}$ & $\mathbf{64.9_{1.0}}$\\
    SDC (11.3k) & $42.9_{0.9}$ & $65.3_{0.8}$ & $67.0_{0.6}$ & $83.6_{0.5}$ & $\mathbf{74.4_{0.5}}$ & $\mathbf{88.9_{0.3}}$ & $\mathbf{51.0_{0.5}}$ & $62.8_{0.6}$\\
    \midrule
    O + F (34.4k) & $\mathbf{49.5_{0.5}}$ & $\mathbf{71.1_{0.6}}$ & $61.6_{0.8}$ & $79.5_{0.6}$ & $58.3_{2.0}$ & $78.5_{1.2}$ & $72.6_{0.4}$ & $80.0_{0.4}$\\
    O + R  (34.4k) & $47.6_{0.7}$ & $69.5_{0.5}$ & $\mathbf{69.2_{0.5}}$ & $\mathbf{84.6_{0.5}}$ & $71.1_{0.7}$ & $86.8_{0.3}$ & $72.8_{0.6}$ & $80.3_{0.5}$\\
    O + SDC  (34.4k) & $41.5_{0.4}$ & $64.2_{0.4}$ & $67.3_{0.6}$ & $84.3_{0.4}$ & $\mathbf{75.0_{0.6}}$ & $\mathbf{88.9_{0.2}}$ & $73.0_{0.2}$ & $80.4_{0.1}$\\
    \midrule
    \multicolumn{9}{c}{Finetuned model: ELECTRA\textsubscript{large}}\\
    \midrule
    Original (O; 23.1k) & $7.5$ & $17.1$ & $29.6$ & $45.2$ & $32.5$ & $49.1$ & $74.2$ & $81.2$\\
    Original (11.3k) & $8.4_{0.9}$ & $18.7_{0.6}$ & $28.8_{0.5}$ & $42.7_{0.9}$ & $33.1_{0.7}$ & $48.6_{1.1}$ & $71.8_{0.1}$ & $79.6_{0.1}$ \\
    \midrule
    BERT\textsubscript{fooled} (F; 11.3k) & $40.2_{4.6}$ & $63.4_{3.2}$ & $50.7_{4.7}$ & $68.5_{4.8}$ & $56.1_{4.4}$ & $75.6_{3.0}$ & $41.0_{4.8}$ & $56.6_{4.2}$\\
    BERT\textsubscript{random} (R; 11.3k) & $42.1_{2.7}$ & $63.5_{2.1}$ & $58.8_{2.2}$ & $76.0_{1.5}$ & $65.8_{1.9}$ & $81.7_{1.3}$ & $52.6_{1.9}$ & $67.5_{1.4}$\\
    SDC (11.3k) & $39.2_{0.3}$ & $40.3_{0.4}$ & $59.6_{0.7}$ & $76.1_{0.6}$ & $\mathbf{69.3_{0.7}}$ & $\mathbf{84.2_{0.5}}$ & $\mathbf{55.7_{0.7}}$ & $\mathbf{69.5_{0.5}}$\\
    \midrule
    O + F (34.4k) & $40.9_{3.4}$ & $63.7_{2.3}$ & $52.6_{2.5}$ & $70.8_{2.1}$ & $55.4_{4.5}$ & $74.4_{4.1}$ & $72.7_{1.2}$ & $80.5_{1.0}$\\
    O + R  (34.4k) & $41.5_{5.6}$ & $61.9_{5.7}$ & $58.6_{4.6}$ & $75.0_{4.4}$ & $64.4_{4.1}$ & $80.4_{3.3}$ & $72.6_{2.0}$ & $80.3_{2.1}$\\
    O + SDC  (34.4k) & $38.0_{0.6}$ & $58.7_{0.6}$ & $59.4_{0.6}$ & $76.1_{0.4}$ & $\mathbf{70.9_{0.4}}$ & $\mathbf{85.1_{0.3}}$ & $73.6_{0.7}$ & $81.2_{0.4}$\\
\bottomrule
  \end{tabular}
  \caption{EM and F1 scores of various models evaluated on adversarial and non-adversarial datasets. Adversarial results in bold are statistically significant compared to SDC setting and vice versa with $p<0.05.$
  \label{tab:ts_collected_bert}\vspace{-4.5mm}}
\end{table*}

\endgroup

\begin{table*}[th]
  \centering
  \tiny
  \begin{tabular}{ l c c c c c c}
    \toprule
    Evaluation set $\rightarrow$ & \multicolumn{2}{c}{D\textsubscript{RoBERTa}} & \multicolumn{2}{c}{D\textsubscript{BERT}} & \multicolumn{2}{c}{D\textsubscript{BiDAF}} \\
    Training set $\downarrow$ & EM & F1 & EM & F1 & EM & F1\\
     \midrule
        \multicolumn{7}{c}{Finetuned model: BERT\textsubscript{large}}\\
    \midrule
    Original (23.1k) & $6.0$ & $13.5$ & $8.1$ & $14.2$ & $12.6$ & $21.4$\\
    Original (11.3k) & $5.4_{0.3}$ & $12.2_{0.1}$ & $7.0_{0.6}$ & $13.6_{0.8}$ & $11.0_{0.9}$ & $19.4_{0.7}$ \\

    \midrule
    
    BERT\textsubscript{fooled} (11.3k) & $11.0_{2.6}$ & $21.0_{3.0}$ & $14.6_{3.7}$ & $24.7_{4.0}$ & $25.1_{6.5}$ & $39.1_{6.9}$\\
    BERT\textsubscript{random} (11.3k) & $\mathbf{12.4_{1.6}}$ & $22.1_{2.2}$ & $16.4_{3.0}$ & $26.2_{2.7}$ & $29.6_{3.7}$ & $43.7_{4.0}$\\
    SDC (11.3k) & $9.1_{0.7}$ & $20.4_{0.7}$ & $14.0_{1.0}$ & $24.6_{0.7}$ & $30.1_{1.2}$ & $43.8_{1.2}$\\
    
    \midrule
    
    Orig + BERT\textsubscript{fooled} (34.4k) & $15.2_{0.8}$ & $25.1_{0.6}$ & $20.4_{0.4}$ & $31.0_{0.4}$ & $32.4_{0.6}$ & $47.0_{0.6}$\\
    Orig + BERT\textsubscript{random}  (34.4k) & $\mathbf{16.9_{0.5}}$ & $\mathbf{23.9_{0.5}}$ & $\mathbf{20.5_{0.6}}$ & $\mathbf{31.2_{0.9}}$ & $\mathbf{34.1_{0.4}}$ & $47.8_{0.7}$\\
    Orig + SDC  (34.4k) & $9.4_{0.6}$ & $20.2_{0.5}$ & $15.3_{1.0}$ & $25.8_{1.1}$ & $32.7_{1.2}$ & $47.2_{1.0}$\\

    \midrule
    
    \multicolumn{7}{c}{Finetuned model: RoBERTa\textsubscript{large}}\\
    \midrule
    Original (23.1k) & $15.7$ & $25.0$ & $26.5$ & $37.0$ & $37.9$ & $50.4$\\
    Original (11.3k) & $14.6_{0.3}$ & $23.8_{0.5}$ & $22.5_{1.2}$ & $32.6_{1.5}$ & $36.0_{1.1}$ & $48.9_{1.2}$ \\

    \midrule
    
    BERT\textsubscript{fooled} (11.3k) & $\mathbf{21.9_{1.6}}$ & $\mathbf{32.2_{1.6}}$ & $30.2_{1.6}$ & $42.5_{1.6}$ & $46.3_{1.6}$ & $61.9_{1.5}$\\
    BERT\textsubscript{random} (11.3k) & $21.3_{1.3}$ & $31.6_{1.5}$ & $\mathbf{31.3_{2.2}}$ & $\mathbf{43.6_{2.3}}$ & $\mathbf{48.0_{1.4}}$ & $\mathbf{63.4_{1.3}}$\\
    SDC (11.3k) & $12.8_{1.2}$ & $23.4_{1.3}$ & $20.0_{1.8}$ & $32.1_{2.2}$ & $40.0_{2.0}$ & $55.0_{1.8}$\\
    
    \midrule
    
    Orig + BERT\textsubscript{fooled} (34.4k) & $\mathbf{25.2_{0.9}}$ & $\mathbf{36.4_{1.0}}$ & $\mathbf{35.9_{0.9}}$ & $\mathbf{48.5_{0.8}}$ & $49.6_{0.7}$ & $65.1_{1.1}$\\
    Orig + BERT\textsubscript{random}  (34.4k) & $24.6_{1.5}$ & $35.2_{1.5}$ & $35.7_{1.0}$ & $48.0_{1.2}$ & $\mathbf{50.6_{1.5}}$ & $\mathbf{65.8_{1.2}}$\\
    Orig + SDC  (34.4k) & $16.1_{0.8}$ & $27.6_{1.1}$ & $26.6_{0.8}$ & $39.7_{0.6}$ & $43.4_{0.4}$ & $59.4_{0.3}$\\

    \midrule
    
    \multicolumn{7}{c}{Finetuned model: ELECTRA\textsubscript{large}}\\
    \midrule
    Original (23.1k) & $8.2$ & $17.4$ & $15.7$ & $24.2$ & $22.4$ & $34.3$\\
    Original (11.3k) & $8.5_{0.4}$ & $16.7_{0.5}$ & $14.3_{1.0}$ & $23.0_{0.9}$ & $20.7_{1.4}$ & $32.0_{1.3}$ \\

    \midrule
    
    BERT\textsubscript{fooled} (11.3k) & $13.8_{3.7}$ & $24.3_{5.6}$ & $18.8_{6.0}$ & $31.1_{8.1}$ & $29.1_{9.0}$ & $44.3_{11.0}$\\
    BERT\textsubscript{random} (11.3k) & $\mathbf{14.8_{1.8}}$ & $\mathbf{25.9_{1.1}}$ & $\mathbf{22.3_{2.9}}$ & $\mathbf{34.6_{2.5}}$ & $34.8_{3.4}$ & $50.5_{2.7}$\\
    SDC (11.3k) & $11.6_{0.6}$ & $22.7_{0.7}$ & $17.8_{1.2}$ & $30.4_{1.3}$ & $32.5_{1.8}$ & $49.3_{1.6}$\\
    
    \midrule
    
    Orig + BERT\textsubscript{fooled} (34.4k) & $16.5_{3.8}$ & $28.0_{3.8}$ & $23.1_{3.9}$ & $35.6_{4.2}$ & $34.8_{5.1}$ & $50.2_{5.7}$\\
    Orig + BERT\textsubscript{random}  (34.4k) & $18.4_{4.2}$ & $28.9_{5.0}$ & $25.9_{5.9}$ & $37.2_{6.9}$ & $37.2_{7.5}$ & $51.1_{9.1}$\\
    Orig + SDC  (34.4k) & $15.6_{1.1}$ & $27.0_{1.1}$ & $22.7_{0.6}$ & $36.0_{0.8}$ & $34.5_{0.9}$ & $49.5_{1.2}$\\

\bottomrule
  \end{tabular}
   \caption{EM and F1 scores of various models evaluated on dev datasets of \citet{bartolo2020beat}. Adversarial results in bold are statistically significant compared to SDC setting and vice versa with $p<0.05.$
  \label{tab:ts_beattheai_bert}
  \vspace{-3.5mm}}
\end{table*}

\begingroup
\setlength{\tabcolsep}{3.35pt}
\begin{table*}[!ht]
  \centering
  \tiny
  \begin{tabular}{ l c c c c c c c c c c c c}
    \toprule
    \multicolumn{13}{c}{Finetuned model: BERT\textsubscript{large}}\\
    \midrule
    Evaluation set $\rightarrow$ & \multicolumn{2}{c}{BioASQ} & \multicolumn{2}{c}{DROP} & \multicolumn{2}{c}{DuoRC} & \multicolumn{2}{c}{Relation Extraction} & \multicolumn{2}{c}{RACE} & \multicolumn{2}{c}{TextbookQA}\\
    Training set $\downarrow$ & EM & F1 & EM & F1 & EM & F1 & EM & F1 & EM & F1 & EM & F1\\
     \midrule
    Original (23.1k) & $19.4$ & $32.5$ & $7.8$ & $16.2$ & $14.5$ & $22.8$ & $32.0$ & $47.1$ & $11.4$ & $18.8$ & $25.0$ & $33.4$\\
    Original (11.3k) & $20.8_{1.7}$ & $36.0_{3.4}$ & $6.2_{1.4}$ & $12.7_{1.8}$ & $13.1_{1.1}$ & $19.8_{1.6}$ & $42.4_{0.4}$ & $55.9_{0.1}$ & $10.3_{0.6}$ & $18.3_{0.4}$ & $20.0_{0.9}$ & $27.9_{0.7}$\\
    \midrule
    BERT\textsubscript{fooled} (11.3k) & $23.5_{6.0}$ & $30.3_{3.5}$ & $11.5_{3.2}$ & $22.2_{3.4}$ & $20.3_{4.5}$ & $28.2_{5.0}$ & $51.5_{8.2}$ & $68.9_{6.6}$ & $15.1_{3.1}$ & $26.1_{4.3}$ & $16.7_{3.8}$ & $24.7_{4.6}$\\
    BERT\textsubscript{random} (11.3k) & $30.3_{3.5}$ & $46.8_{2.8}$ & $14.4_{2.0}$ & $25.1_{2.5}$ & $26.7_{3.3}$ & $35.3_{3.0}$ & $61.3_{5.8}$ & $75.9_{4.5}$ & $18.4_{1.8}$ & $29.9_{2.0}$ & $21.9_{3.1}$ & $30.9_{3.8}$\\
    SDC (11.3k) & $\mathbf{35.1_{2.1}}$ & $\mathbf{55.7_{1.1}}$ & $14.6_{0.4}$ & $24.7_{0.6}$ & $\mathbf{31.7_{0.7}}$ & $\mathbf{41.2_{0.7}}$ & $63.2_{1.2}$ & $77.7_{0.7}$ & $\mathbf{19.7_{0.6}}$ & $31.0_{0.6}$ & $\mathbf{26.0_{4.3}}$ & $\mathbf{35.5_{4.7}}$\\
    \midrule
    Orig + Fooled (34.4k) & $31.7_{1.2}$ & $48.2_{1.2}$ & $19.9_{0.9}$ & $31.0_{0.8}$ & $24.4_{0.9}$ & $33.1_{1.4}$ & $55.0_{1.7}$ & $71.5_{1.2}$ & $19.2_{1.3}$ & $31.0_{1.1}$ & $22.2_{4.7}$ & $30.9_{5.4}$\\
    Orig + Random  (34.4k) & $34.9_{1.2}$ & $51.8_{0.9}$ & $\mathbf{21.4_{0.6}}$ & $\mathbf{33.1_{0.4}}$ & $27.1_{1.2}$ & $36.1_{1.2}$ & $62.3_{0.9}$ & $77.1_{0.7}$ & $21.0_{1.4}$ & $33.0_{1.3}$ & $27.7_{3.9}$ & $37.1_{4.0}$\\
    Orig + SDC  (34.4k) & $\mathbf{38.8_{1.5}}$ & $\mathbf{56.0_{1.3}}$ & $19.4_{0.9}$ & $31.1_{1.0}$ & $\mathbf{31.9_{0.4}}$ & $\mathbf{41.6_{0.6}}$ & $62.4_{0.7}$ & $77.8_{0.2}$ & $20.7_{1.4}$ & $32.7_{1.2}$ & $29.0_{2.4}$ & $38.8_{3.1}$\\
    \midrule
    & \multicolumn{2}{c}{HotpotQA}  & \multicolumn{2}{c}{Natural Questions} & \multicolumn{2}{c}{NewsQA} & \multicolumn{2}{c}{SearchQA} & \multicolumn{2}{c}{SQuAD} & \multicolumn{2}{c}{TriviaQA}\\
    & EM & F1 & EM & F1 & EM & F1 & EM & F1 & EM & F1 & EM & F1\\
     \midrule
    Original (23.1k) & $19.4$ & $33.9$ & $36.3$ & $48.7$ & $16.2$ & $25.6$ & $11.3$ & $19.3$ & $32.5$ & $46.0$ & $16.8$ & $25.3$ \\
    Original (11.3k) & $20.1_{0.3}$ & $32.6_{0.6}$ & $38.4_{0.5}$ & $50.6_{0.6}$ & $15.0_{1.0}$ & $24.9_{1.7}$ & $11.1_{0.7}$ & $18.6_{1.2}$ & $29.6_{0.4}$ & $43.0_{0.7}$  & $15.3_{1.0}$ & $23.9_{1.4}$ \\
     \midrule
    BERT\textsubscript{fooled} (11.3k) & $27.2_{6.4}$ & $43.2_{7.5}$ & $28.0_{5.7}$ & $42.8_{6.5}$ & $22.7_{4.7}$ & $37.5_{6.4}$ & $6.1_{1.7}$ & $11.8_{2.2}$ & $42.6_{7.6}$ & $60.6_{7.9}$ & $16.1_{4.6}$ & $24.3_{5.4}$\\
    BERT\textsubscript{random} (11.3k) & $37.5_{3.1}$ & $54.4_{3.1}$ & $36.7_{3.9}$ & $51.2_{3.5}$ & $29.6_{1.9}$ & $44.9_{1.9}$ & $8.6_{1.4}$ & $14.6_{1.8}$ & $51.9_{2.6}$ & $69.3_{2.1}$ & $24.7_{2.8}$ & $34.4_{3.0}$\\
    SDC (11.3k) & $\mathbf{41.2_{0.9}}$ & $\mathbf{57.9_{1.0}}$ & $39.3_{1.2}$ & $53.6_{1.1}$ & $\mathbf{32.0_{0.8}}$ & $\mathbf{48.0_{1.1}}$ & $\mathbf{10.6_{1.4}}$ & $\mathbf{18.0_{1.3}}$ & $\mathbf{56.4_{0.4}}$ & $\mathbf{72.5_{0.4}}$ & $\mathbf{28.6_{0.8}}$ & $\mathbf{39.9_{0.9}}$\\
    \midrule
    Orig + Fooled (34.4k) & $34.4_{1.0}$ & $51.1_{0.8}$ & $39.9_{1.3}$ & $54.1_{0.8}$ & $26.3_{0.9}$ & $42.8_{1.1}$ & $8.7_{1.5}$ & $14.5_{1.7}$ & $47.6_{0.5}$ & $66.3_{0.5}$ & $21.9_{0.7}$ & $30.9_{0.8}$\\
    Orig + Random (34.4k) & $41.0_{0.7}$ & $57.3_{0.7}$ & $44.5_{0.4}$ & $58.2_{0.2}$ & $30.0_{0.5}$ & $45.9_{0.6}$ & $11.2_{0.7}$ & $17.7_{0.9}$ & $53.4_{0.4}$ & $70.8_{0.4}$ & $28.6_{1.3}$ & $38.6_{1.4}$\\
    Orig + SDC (34.4k) & $\mathbf{43.3_{0.2}}$ & $\mathbf{60.0_{0.3}}$ & $45.6_{0.9}$ & $58.7_{1.1}$ & $\mathbf{32.0_{0.8}}$ & $\mathbf{48.6_{1.1}}$ & $\mathbf{13.6_{0.4}}$ & $\mathbf{22.2_{0.5}}$ & $\mathbf{57.0_{0.3}}$ & $\mathbf{73.2_{0.3}}$ & $\mathbf{30.9_{1.0}}$ & $\mathbf{42.4_{0.9}}$\\
    \midrule
    \multicolumn{13}{c}{Finetuned model: RoBERTa\textsubscript{large}}\\
    \midrule
    Evaluation set $\rightarrow$ & \multicolumn{2}{c}{BioASQ} & \multicolumn{2}{c}{DROP} & \multicolumn{2}{c}{DuoRC} & \multicolumn{2}{c}{Relation Extraction} & \multicolumn{2}{c}{RACE} & \multicolumn{2}{c}{TextbookQA}\\
    Training set $\downarrow$ & EM & F1 & EM & F1 & EM & F1 & EM & F1 & EM & F1 & EM & F1\\
     \midrule
    Original (23.1k) & $47.7$ & $63.5$ & $37.2$ & $48.1$ & $38.6$ & $49.1$ & $74.4$ & $85.9$ & $33.7$ & $44.9$ & $36.4$ & $46$ \\
    Original (11.3k) & $46.3_{0.1}$ & $62.7_{1.0}$ & $34.7_{0.3}$ & $46.5_{0.8}$ & $36.6_{1.8}$ & $46.9_{2.1}$ & $72.3_{0.8}$ & $84.5_{0.3}$ & $30.7_{0.2}$ & $42.2_{0.3}$ & $34.9_{0.4}$ & $44.4_{0.2}$ \\
     \midrule
    BERT\textsubscript{fooled} (11.3k) & $35.6_{1.3}$ & $51.0_{1.2}$ & $34.1_{2.5}$ & $46.8_{2.4}$ & $31.4_{2.5}$ & $39.7_{3.0}$ & $67.0_{1.0}$ & $81.9_{0.5}$ & $28.2_{1.3}$ & $41.4_{1.1}$ & $25.4_{2.4}$ & $35.1_{2.4}$\\
    BERT\textsubscript{random} (11.3k) & $40.4_{1.2}$ & $57.4_{1.2}$ & $\mathbf{38.1_{2.2}}$ & $\mathbf{51.2_{2.0}}$ & $36.7_{1.6}$ & $45.5_{1.7}$ & $71.0_{0.5}$ & $84.4_{0.3}$ & $\mathbf{31.6_{1.3}}$ & $\mathbf{45.3_{1.1}}$ & $29.8_{1.4}$ & $39.3_{1.6}$\\
    SDC (11.3k) & $\mathbf{41.3_{1.0}}$ & $\mathbf{59.7_{1.0}}$ & $24.4_{2.2}$ & $38.9_{2.9}$ & $\mathbf{41.1_{0.8}}$ & $\mathbf{51.8_{0.5}}$ & $\mathbf{72.6_{0.6}}$ & $84.6_{0.3}$ & $29.5_{1.1}$ & $43.3_{1.2}$ & $\mathbf{35.6_{1.8}}$ & $\mathbf{46.1_{1.7}}$\\
    \midrule
    Orig + Fooled (34.4k) & $41.2_{1.2}$ & $56.7_{0.9}$ & $43.3_{1.4}$ & $54.7_{1.6}$ & $32.0_{0.7}$ & $41.5_{1.0}$ & $61.3_{2.3}$ & $78.3_{1.2}$ & $31.7_{0.6}$ & $45.7_{1.0}$ & $37.6_{2.5}$ & $48.0_{2.6}$\\
    Orig + Random (34.4k) & $\mathbf{45.7_{1.0}}$ & $\mathbf{62.2_{0.8}}$ & $\mathbf{46.5_{1.4}}$ & $\mathbf{58.0_{1.2}}$ & $38.9_{0.9}$ & $48.9_{0.8}$ & $67.6_{1.2}$ & $82.6_{0.9}$ & $33.6_{1.1}$ & $\mathbf{47.1_{0.7}}$ & $40.0_{1.6}$ & $50.3_{1.7}$\\
    Orig + SDC (34.4k) & $43.1_{0.8}$ & $60.9_{0.4}$ & $40.2_{1.4}$ & $53.8_{0.8}$ & $\mathbf{40.0_{1.4}}$ & $\mathbf{51.9_{1.5}}$ & $\mathbf{70.9_{0.4}}$ & $\mathbf{83.3_{0.4}}$ & $32.9_{0.8}$ & $45.7_{0.7}$ & $40.9_{1.1}$ & $51.9_{1.3}$\\
    \midrule
    & \multicolumn{2}{c}{HotpotQA}  & \multicolumn{2}{c}{Natural Questions} & \multicolumn{2}{c}{NewsQA} & \multicolumn{2}{c}{SearchQA} & \multicolumn{2}{c}{SQuAD} & \multicolumn{2}{c}{TriviaQA}\\
    & EM & F1 & EM & F1 & EM & F1 & EM & F1 & EM & F1 & EM & F1\\
     \midrule
    Original (23.1k) & $48.1$ & $63.5$ & $55.3$ & $67.6$ & $38.6$ & $54.4$ & $39.7$ & $49.3$ & $61.9$ & $76.7$ & $47.5$ & $59.6$ \\
    Original (11.3k) & $46.6_{0.3}$ & $63.2_{0.3}$ & $54.6_{0.4}$ & $66.9_{0.4}$ & $36.3_{1.0}$ & $51.6_{1.2}$ & $33.8_{0.8}$ & $43.0_{0.6}$ & $60.1_{0.4}$ & $75.3_{0.3}$ & $44.9_{0.6}$ & $57.2_{0.7}$\\
     \midrule
    BERT\textsubscript{fooled} (11.3k) & $46.5_{0.8}$ & $63.3_{0.8}$ & $41.6_{1.2}$ & $56.6_{1.1}$ & $33.8_{1.2}$ & $50.7_{1.6}$ & $15.3_{1.9}$ & $21.5_{1.9}$ & $60.0_{0.6}$ & $77.6_{0.5}$ & $37.0_{1.7}$ & $45.9_{2.1}$\\
    BERT\textsubscript{random} (11.3k) & $50.7_{0.6}$ & $67.7_{0.7}$ & $48.1_{0.9}$ & $62.6_{0.8}$ & $39.5_{0.8}$ & $56.1_{1.1}$ & $17.0_{1.7}$ & $23.6_{1.8}$ & $65.4_{0.4}$ & $81.4_{0.3}$ & $43.3_{1.1}$ & $52.5_{1.2}$\\
    SDC (11.3k) & $\mathbf{52.0_{1.3}}$ & $68.7_{1.4}$ & $47.9_{1.2}$ & $61.7_{1.3}$ & $\mathbf{44.0_{0.9}}$ & $\mathbf{61.9_{0.7}}$ & $\mathbf{24.9_{2.0}}$ & $\mathbf{33.0_{2.0}}$ & $\mathbf{66.4_{0.6}}$ & $\mathbf{82.2_{0.5}}$ & $\mathbf{47.0_{0.6}}$ & $\mathbf{58.3_{0.7}}$\\
    \midrule
    Orig + Fooled (34.4k) & $47.2_{1.1}$ & $64.7_{1.1}$ & $53.2_{0.7}$ & $66.8_{0.6}$ & $33.9_{0.7}$ & $52.0_{0.7}$ & $28.2_{2.1}$ & $35.3_{2.5}$ & $58.2_{0.8}$ & $76.9_{0.6}$ & $38.8_{0.9}$ & $48.6_{1.0}$\\
    Orig + Random (34.4k) & $53.2_{0.5}$ & $70.1_{0.5}$ & $54.8_{0.4}$ & $68.2_{0.3}$ & $41.6_{0.6}$ & $58.9_{0.7}$ & $30.6_{1.9}$ & $38.3_{2.0}$ & $65.3_{0.5}$ & $81.8_{0.3}$ & $46.7_{1.0}$ & $57.1_{0.9}$\\
    Orig + SDC (34.4k) & $53.9_{0.9}$ & $70.7_{0.9}$ & $\mathbf{55.9_{0.4}}$ & $\mathbf{68.7_{0.5}}$ & $\mathbf{44.2_{0.3}}$ & $\mathbf{62.5_{0.4}}$ & $\mathbf{36.0_{1.3}}$ & $\mathbf{45.2_{1.6}}$ & $\mathbf{66.6_{0.4}}$ & $\mathbf{82.7_{0.2}}$ & $\mathbf{48.0_{0.8}}$ & $\mathbf{59.8_{0.7}}$\\
    \midrule
    \multicolumn{13}{c}{Finetuned model: ELECTRA\textsubscript{large}}\\
    \midrule
    Evaluation set $\rightarrow$ & \multicolumn{2}{c}{BioASQ} & \multicolumn{2}{c}{DROP} & \multicolumn{2}{c}{DuoRC} & \multicolumn{2}{c}{Relation Extraction} & \multicolumn{2}{c}{RACE} & \multicolumn{2}{c}{TextbookQA}\\
    Training set $\downarrow$ & EM & F1 & EM & F1 & EM & F1 & EM & F1 & EM & F1 & EM & F1\\
     \midrule
    Original (23.1k) & $29.1$ & $42.8$ & $17.6$ & $26.9$ & $18.9$ & $27.1$ & $53.4$ & $67.4$ & $19.6$ & $28.5$ & $32.5$ & $41.8$ \\
    Original (11.3k) & $33.1_{1.4}$ & $49.4_{2.5}$ & $15.5_{1.8}$ & $26.5_{1.1}$ & $21.2_{0.8}$ & $29.4_{0.6}$ & $54.9_{0.9}$ & $69.4_{1.1}$ & $18.0_{0.8}$ & $28.4_{0.7}$ & $29.2_{0.5}$ & $37.8_{0.3}$ \\
     \midrule
    BERT\textsubscript{fooled} (11.3k) & $32.4_{4.6}$ & $50.2_{3.6}$ & $19.9_{4.3}$ & $33.4_{3.5}$ & $25.2_{4.2}$ & $35.1_{3.7}$ & $57.0_{4.9}$ & $74.6_{3.1}$ & $20.6_{2.5}$ & $34.0_{2.5}$ & $19.5_{3.3}$ & $28.5_{4.0}$\\
    BERT\textsubscript{random} (11.3k) & $37.1_{2.9}$ & $55.1_{2.1}$ & $\mathbf{21.1_{1.9}}$ & $\mathbf{35.0_{1.6}}$ & $30.5_{2.1}$ & $40.3_{1.6}$ & $64.3_{2.9}$ & $78.7_{1.3}$ & $23.3_{1.5}$ & $36.5_{1.5}$ & $25.7_{3.3}$ & $35.1_{3.5}$\\
    SDC (11.3k) & $\mathbf{40.6_{1.7}}$ & $\mathbf{59.2_{1.4}}$ & $17.5_{0.9}$ & $30.7_{1.1}$ & $\mathbf{33.3_{2.1}}$ & $\mathbf{43.6_{1.9}}$ & $65.9_{1.4}$ & $79.6_{0.8}$ & $23.4_{1.1}$ & $35.5_{1.0}$ & $27.4_{2.7}$ & $36.8_{2.9}$\\
    \midrule
    Orig + Fooled (34.4k) & $31.7_{1.3}$ & $48.2_{1.3}$ & $19.9_{0.9}$ & $31.0_{0.8}$ & $24.5_{0.9}$ & $33.1_{1.4}$ & $55.0_{1.7}$ & $71.5_{1.2}$ & $19.2_{1.3}$ & $31.0_{1.1}$ & $22.2_{4.7}$ & $30.9_{5.4}$\\
    Orig + Random (34.4k) & $37.8_{5.2}$ & $54.4_{5.4}$ & $\mathbf{27.6_{6.8}}$ & $\mathbf{39.4_{8.1}}$ & $28.4_{5.1}$ & $38.2_{5.7}$ & $62.9_{6.8}$ & $77.2_{5.2}$ & $\mathbf{24.3_{4.6}}$ & $\mathbf{37.4_{5.3}}$ & $\mathbf{34.0_{6.1}}$ & $\mathbf{43.5_{6.2}}$\\
    Orig + SDC (34.4k) & $\mathbf{40.0_{0.9}}$ & $\mathbf{57.6_{0.9}}$ & $19.4_{0.9}$ & $31.1_{1.0}$ & $\mathbf{31.9_{0.4}}$ & $\mathbf{41.6_{0.6}}$ & $62.4_{0.7}$ & $76.8_{0.2}$ & $19.5_{1.4}$ & $31.7_{1.2}$ & $29.0_{2.4}$ & $38.8_{3.1}$\\
    \midrule
    & \multicolumn{2}{c}{HotpotQA}  & \multicolumn{2}{c}{Natural Questions} & \multicolumn{2}{c}{NewsQA} & \multicolumn{2}{c}{SearchQA} & \multicolumn{2}{c}{SQuAD} & \multicolumn{2}{c}{TriviaQA}\\
    & EM & F1 & EM & F1 & EM & F1 & EM & F1 & EM & F1 & EM & F1\\
     \midrule
    Original (23.1k) & $29.6$ & $43$ & $40.9$ & $55.3$ & $20.4$ & $32.2$ & $21.5$ & $30.3$ & $39.9$ & $54.8$ & $21$ & $31.2$ \\
    Original (11.3k) & $26.8_{0.2}$ & $39.7_{0.2}$ & $38.7_{0.9}$ & $54.2_{0.9}$ & $21.0_{1.0}$ & $33.2_{1.1}$ & $17.2_{1.5}$ & $24.8_{1.6}$ & $40.5_{1.2}$ & $55.9_{1.2}$ & $23.9_{1.8}$ & $33.5_{1.8}$ \\
     \midrule
    BERT\textsubscript{fooled} (11.3k) & $36.7_{4.0}$ & $54.2_{2.9}$ & $35.1_{3.8}$ & $51.7_{3.1}$ & $28.5_{2.4}$ & $45.1_{2.4}$ & $7.0_{1.3}$ & $13.9_{1.7}$ & $48.3_{4.2}$ & $67.5_{3.4}$ & $23.8_{2.9}$ & $34.5_{2.3}$\\
    BERT\textsubscript{random} (11.3k) & $41.4_{2.4}$ & $58.4_{1.6}$ & $43.2_{1.7}$ & $58.5_{1.3}$ & $33.3_{1.6}$ & $49.8_{1.6}$ & $9.2_{1.5}$ & $16.8_{2.1}$ & $55.4_{2.3}$ & $72.9_{1.7}$ & $28.9_{1.4}$ & $39.9_{1.0}$\\
    SDC (11.3k) & $43.0_{1.4}$ & $59.6_{1.1}$ & $\mathbf{46.1_{1.0}}$ & $\mathbf{60.4_{0.8}}$ & $\mathbf{35.3_{1.1}}$ & $\mathbf{51.9_{1.1}}$ & $10.5_{1.4}$ & $\mathbf{19.0_{1.6}}$ & $\mathbf{58.6_{1.4}}$ & $\mathbf{74.9_{1.0}}$ & $29.0_{1.6}$ & $60.7_{1.3}$\\
    \midrule
    Orig + Fooled (34.4k) & $34.4_{1.0}$ & $51.1_{0.8}$ & $45.4_{2.9}$ & $59.9_{2.6}$ & $26.3_{0.9}$ & $42.8_{1.1}$ & $8.7_{1.5}$ & $14.5_{1.7}$ & $47.6_{0.5}$ & $66.3_{0.5}$ & $21.9_{0.7}$ & $30.9_{0.8}$\\
    Orig + Random (34.4k) & $41.4_{4.7}$ & $57.4_{4.5}$ & $46.2_{3.8}$ & $60.0_{3.5}$ & $31.7_{4.2}$ & $47.5_{5.2}$ & $14.9_{2.2}$ & $23.1_{2.2}$ & $55.2_{4.6}$ & $72.1_{4.6}$ & $29.8_{5.2}$ & $40.2_{5.2}$\\
    Orig + SDC (34.4k) & $\mathbf{43.9_{0.5}}$ & $\mathbf{60.4_{0.3}}$ & $49.4_{0.5}$ & $63.0_{0.7}$ & $\mathbf{32.4_{0.7}}$ & $\mathbf{49.0_{0.8}}$ & $13.6_{0.4}$ & $22.2_{0.5}$ & $\mathbf{57.6_{1.0}}$ & $\mathbf{74.0_{1.0}}$ & $\mathbf{31.7_{0.8}}$ & $\mathbf{43.4_{0.6}}$\\
\bottomrule
  \end{tabular}
   \caption{EM and F1 scores of various models evaluated on MRQA dev and test sets. Adversarial results in bold are statistically significant compared to SDC setting and vice versa with $p<0.05.$
  \label{tab:ood_bert_data}}
\end{table*}

\section{Experiments and Results}
Our study allows us to answer three questions: 
(i) how well do models fine-tuned on ADC data 
generalize to unseen distributions 
compared to fine-tuning on SDC? 
(ii) Among the differences between ADC and SDC, how many are due to workers trying to fool the model regardless of whether they are successful?
and (iii) what is the impact of
training on adversarial data only
versus using it as a data augmentation strategy?

\paragraph{Datasets} 
For both BERT and ELECTRA,
we first identify contexts
for which at least one question 
elicited an incorrect model prediction.
Note that this set of contexts
is different for BERT and ELECTRA.
For each such context $c$, we identify 
the number of questions $k_c$ (out of 5)
that successfully fooled the model.
We then create $3$ datasets per model by,
for each context, 
(i) choosing precisely those $k_c$ questions
that fooled the model 
(BERT\textsubscript{fooled} 
and ELECTRA\textsubscript{fooled}); 
(ii) randomly choosing $k_c$ 
questions (out of $5$) from ADC data 
without replacement
(BERT\textsubscript{random} 
and ELECTRA\textsubscript{random})---regardless
of whether they fooled the model; 
and (iii) randomly choosing $k_c$ questions
(out of $5$)
from the SDC data without replacement.
Thus, we create $6$ datasets,
where all $3$ BERT datasets have 
the same number of questions per context
(and $11.3k$ total training examples),
while all $3$ ELECTRA datasets
likewise share the same number of questions per context
(and $14.7k$ total training examples).
See Table \ref{tab:data_stats} 
for details on the number of passages 
and question-answer pairs 
used in the different splits.

\paragraph{Models} 
For our empirical analysis,
we fine-tune BERT \citep{devlin2019bert}, 
RoBERTa \citep{liu2019roberta}, 
and ELECTRA \citep{Clark2020ELECTRA} models 
on all six datasets generated 
as part of our study
(four datasets via ADC: 
BERT\textsubscript{fooled}, 
BERT\textsubscript{random},
ELECTRA\textsubscript{fooled}, 
ELECTRA\textsubscript{random}, 
and the two datasets via SDC). 
We also fine-tune these models after
augmenting the original data to collected datasets.
We report the means and standard deviations (in subscript) 
of EM and F1 scores following 
$10$ runs of each experiment.
Models fine-tuned on all ADC datasets 
typically perform better 
on their held-out test sets 
than those trained on SDC data and vice-versa 
(Table \ref{tab:ts_collected_bert} 
and Appendix Table \ref{tab:ts_collected_electra}).
RoBERTa fine-tuned on the BERT\textsubscript{fooled} training set 
obtains EM and F1 scores of $49.2$ and $71.2$, respectively,
on the BERT\textsubscript{fooled} test set, 
outperforming RoBERTa models 
fine-tuned on BERT\textsubscript{random} 
(EM: $48.0$, F1: $69.8$)
and SDC (EM: $42.0$, F1: $65.3$).
Performance on the original dev set \citep{karpukhin2020dense} 
is generally comparable across all models.

\paragraph{Out-of-domain generalization to adversarial data}
We evaluate these models 
on adversarial test sets constructed 
with BiDAF (D\textsubscript{BiDAF}),  
BERT (D\textsubscript{BERT}) 
and RoBERTa (D\textsubscript{RoBERTa}) in the loop~\citep{bartolo2020beat}. 
Prior work suggests that training on ADC data 
leads to models that perform better 
on similarly constructed 
adversarial evaluation sets.
Both BERT and RoBERTa models 
fine-tuned on adversarial data 
generally outperform models 
fine-tuned on SDC data 
(or when either datasets 
are augmented to the original data) 
on all three evaluation sets 
(Table \ref{tab:ts_beattheai_bert} and 
Appendix Table \ref{tab:ts_beattheai_electra}).
A RoBERTa model fine-tuned on BERT\textsubscript{fooled} 
outperforms a RoBERTa model fine-tuned on SDC 
by~$9.1$,~$9.3$, and~$6.2$ EM points 
on D\textsubscript{RoBERTa}, D\textsubscript{BERT}, 
and D\textsubscript{BiDAF}, respectively.
We observe similar trends on ELECTRA models fine-tuned on ADC data versus SDC data,
but these gains disappear 
when the same models 
are finetuned on augmented data.
For instance, while ELECTRA 
fine-tuned on BERT\textsubscript{random} 
obtains an EM score of~$14.8$ 
on D\textsubscript{RoBERTa}, 
outperforming an ELECTRA fine-tuned 
on SDC data by $\approx 3$ pts, 
the difference is no longer significant 
when respective models are fine-tuned 
after original data is augmented to these datasets. 
ELECTRA models fine-tuned on ADC data 
with ELECTRA in the loop 
perform no better 
than those trained on SDC.
Fine-tuning ELECTRA on SDC augmented to original data 
leads to an $\approx 1$ pt improvement 
on both metrics compared to augmenting ADC.
Overall, we find that models 
fine-tuned on ADC data
typically generalize better 
to out-of-domain adversarial test sets
than models fine-tuned on SDC data, 
confirming the findings by \citet{dinan2019build}.

\paragraph{Out-of-domain generalization to MRQA}
We further evaluate these models 
on $12$ out-of-domain datasets 
used in the 2019 MRQA shared task\footnote{The 
MRQA 2019 shared task includes 
HotpotQA \citep{yang2018hotpotqa}, 
Natural Questions \citep{kwiatkowski2019natural},
SearchQA \citep{dunn2017searchqa}, 
SQuAD \citep{rajpurkar2016squad}, 
TriviaQA \citep{joshi2017triviaqa}, 
BioASQ \citep{tsatsaronis2015overview}, 
DROP \citep{dua2019drop}, 
DuoRC \citep{saha2018duorc}, 
RelationExtraction \citep{levy2017zero}, 
RACE \citep{lai2017race}, 
and TextbookQA \citep{kembhavi2017you}.} 
(Table \ref{tab:ood_bert_data} 
and Appendix Table \ref{tab:ood_electra_data}).\footnote{Interestingly, 
RoBERTa appears to perform better 
compared to BERT and ELECTRA. 
Prior works have hypothesized that 
the bigger size and increased diversity 
of the pre-training corpus of RoBERTa 
(compared to those of BERT and ELECTRA)
might somehow be responsible for
RoBERTa's better out-of-domain generalization, 
\citep{baevski2019cloze, hendrycks2020pretrained, tu2020empirical}.}
Notably, for BERT,  
fine-tuning on SDC data 
leads to significantly better performance 
(as compared to fine-tuning on ADC data collected with BERT)
on $9$ out of $12$ MRQA datasets,
with gains of more than $10$ EM pts on $6$ of them.
On BioASQ, BERT fine-tuned 
on BERT\textsubscript{fooled} 
obtains EM and F1 scores 
of $23.5$ and $30.3$, respectively.
By comparison, fine-tuning on SDC data
yields markedly higher EM and F1 scores 
of $35.1$ and $55.7$, respectively.
Similar trends hold across models and datasets. 
Interestingly, ADC fine-tuning 
often improves performance on DROP compared to SDC.
For instance, RoBERTa fine-tuned 
on ELECTRA\textsubscript{random} 
outperforms RoBERTa fine-tuned on SDC by $\approx7$ pts.
Note that DROP itself was adversarially constructed.
On Natural Questions, models fine-tuned on ADC data
generally perform comparably 
to those fine-tuned on SDC data.
RoBERTa fine-tuned on BERT\textsubscript{random} 
obtains EM and F1 scores of $48.1$ and $62.6$, respectively, 
whereas RoBERTa fine-tuned on SDC data 
obtains scores of $47.9$ and $61.7$, respectively.
It is worth noting that passages sourced
to construct both ADC and SDC datasets 
come from the Natural Questions dataset, 
which could be one reason 
why models fine-tuned on ADC datasets 
perform similar to those fine-tuned on SDC datasets 
when evaluated on Natural Questions.

\paragraph{On the the adversarial process versus adversarial success}
We notice that models 
fine-tuned on BERT\textsubscript{random} 
and ELECTRA\textsubscript{random}
typically outperform models 
fine-tuned on BERT\textsubscript{fooled} 
and ELECTRA\textsubscript{fooled}, respectively,
on adversarial test data collected in prior work \citep{bartolo2020beat}, as well as on MRQA.
Similar observation can be made when the ADC data is augmented with the original training data.
These trends 
suggest that the ADC process 
(regardless of the outcome) 
explains our results 
more than successfully fooling a model.
Furthermore, models 
fine-tuned only on SDC data
tend to  outperform ADC-only fine-tuned models;
however, following augmentation,
ADC fine-tuning achieves comparable performance 
on more datasets than before,
showcasing generalization following augmentation.
Notice that augmenting ADC data 
to original data may not always help. 
BERT fine-tuned on original $23.1k$ examples
achieves an EM $11.3$ on SearchQA. 
When fine-tuned on BERT\textsubscript{fooled} 
augmented to the original data, 
this drops to $8.7$, 
and when fine-tuned on BERT\textsubscript{random} 
augmented to the original data,
it drops to $11.2$.
Fine-tuning on SDC augmented 
to the original data, however, 
results in EM of $13.6$.

\section{Qualitative Analysis}
\begin{figure*}[!t]
\centering
\begin{subfigure}[h]{0.1667\textwidth}
\centering
    \includegraphics[width=\textwidth]{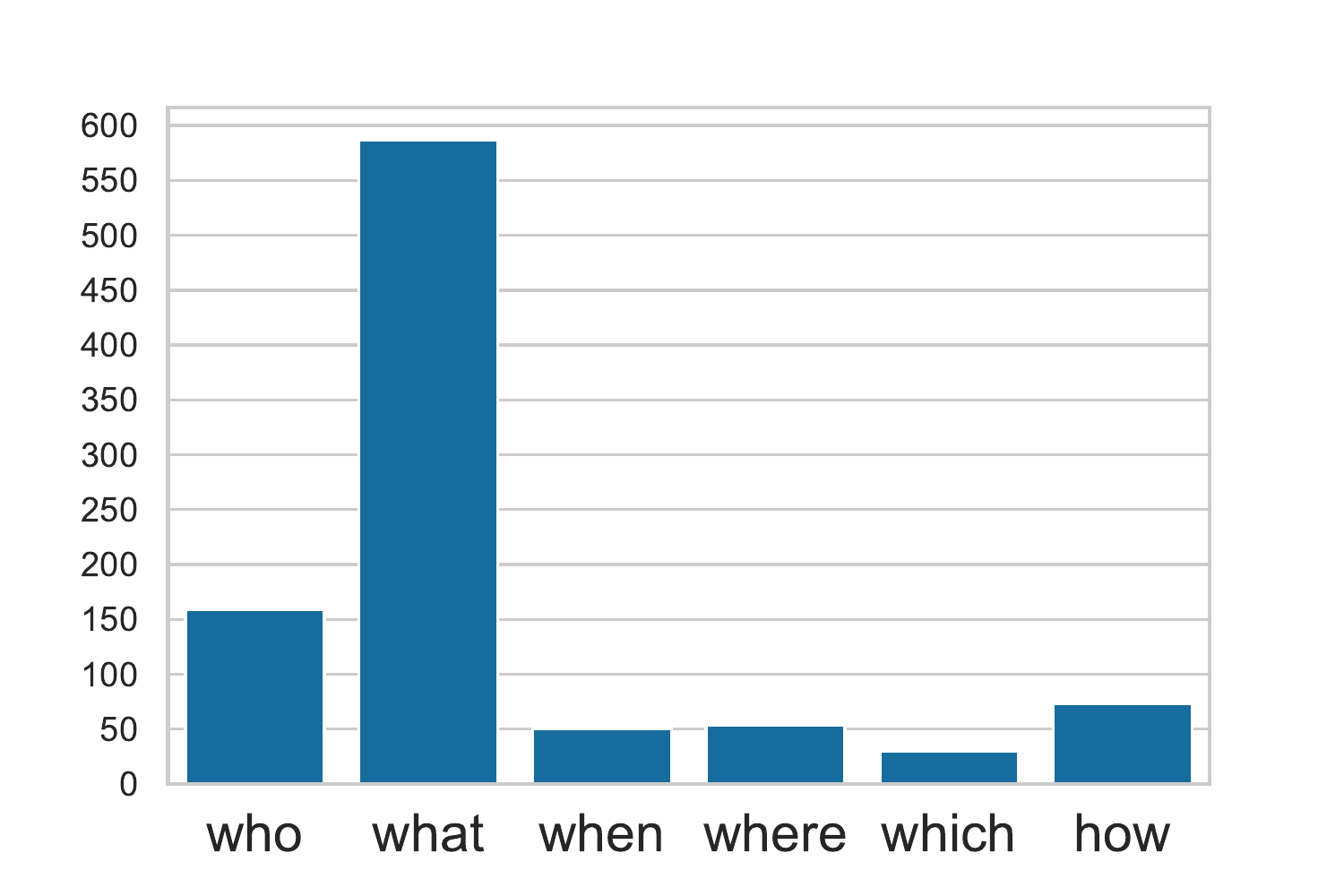}
\caption{
BERT\textsubscript{fooled}
\label{fig:bert_fooled_wh}
}
\end{subfigure}
\hspace{-2.0mm}
\begin{subfigure}[h]{0.1667\textwidth}
\centering
    \includegraphics[width=\textwidth]{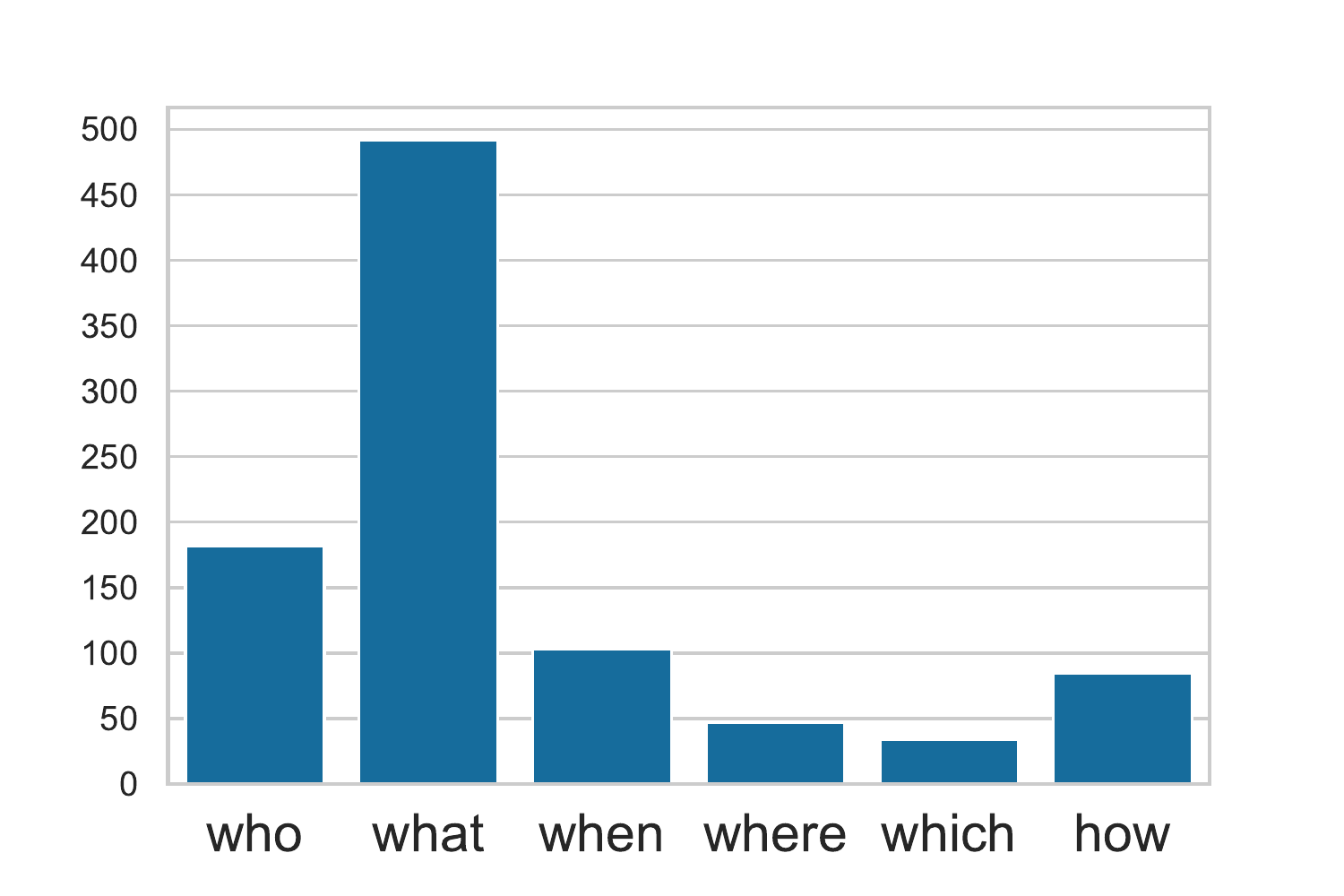}
\caption{
BERT\textsubscript{random}
\label{fig:bert_random_wh}}
\end{subfigure}
\hspace{-2.0mm}
\begin{subfigure}[h]{0.1667\textwidth}
\centering
    \includegraphics[width=\textwidth]{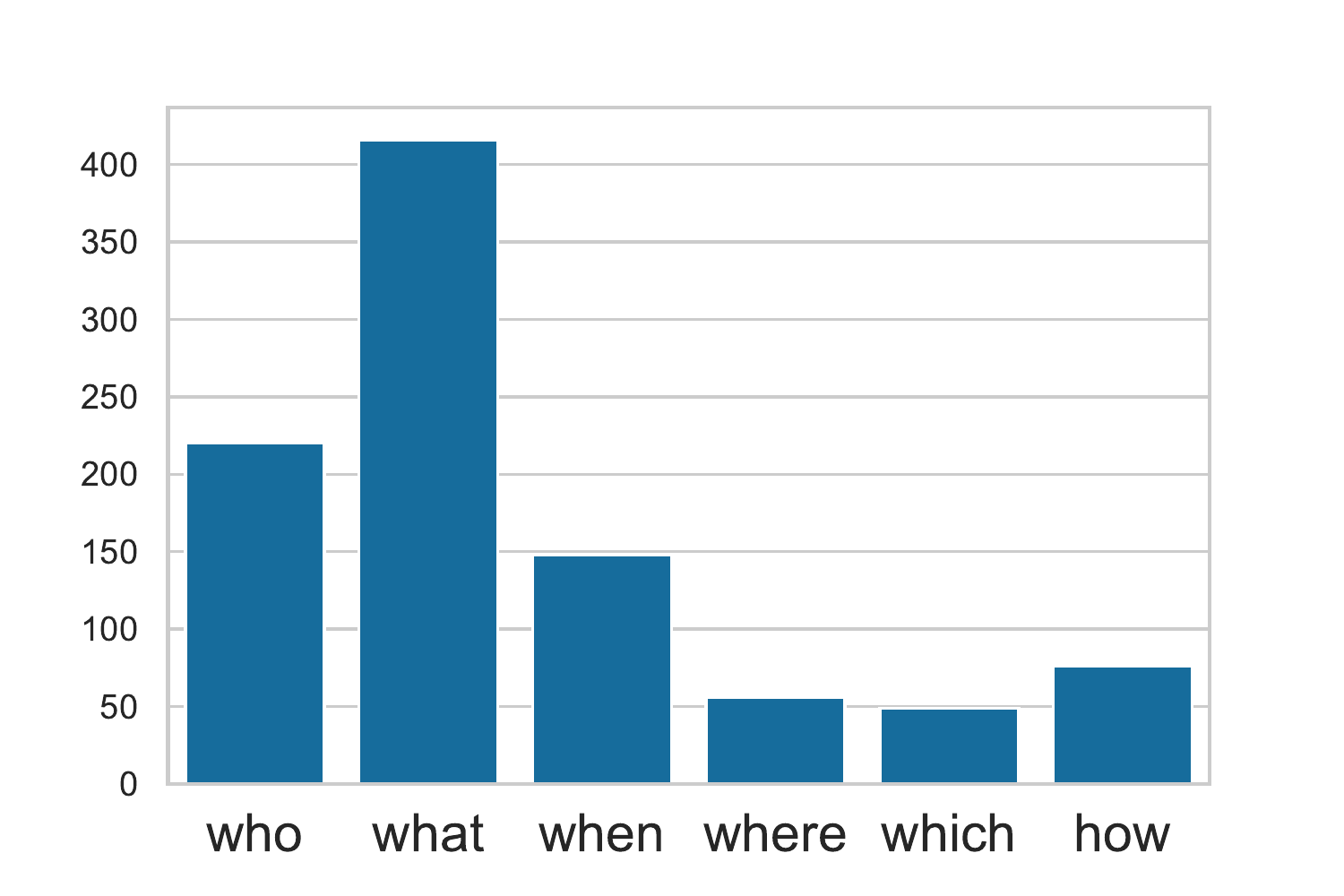}
\caption{
SDC-BERT
\label{fig:bert_nomodel_wh}}
\end{subfigure}
\hspace{-2.0mm}
\begin{subfigure}[h]{0.1667\textwidth}
\centering
    \includegraphics[width=\textwidth]{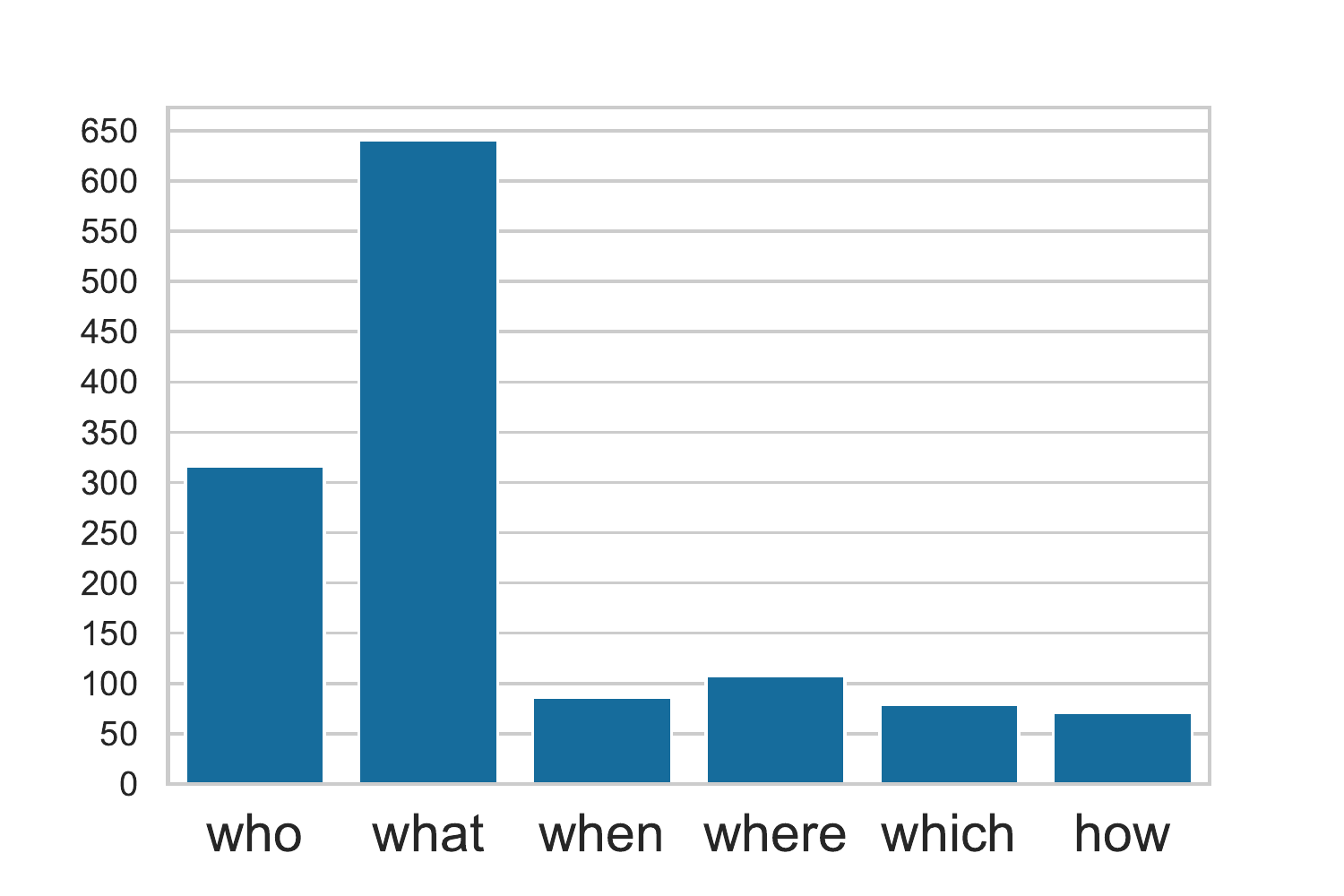}
\caption{
ELECTRA\textsubscript{fooled}
\label{fig:bert_fooled_wh}
}
\end{subfigure}
\begin{subfigure}[h]{0.1667\textwidth}
\centering
    \includegraphics[width=\textwidth]{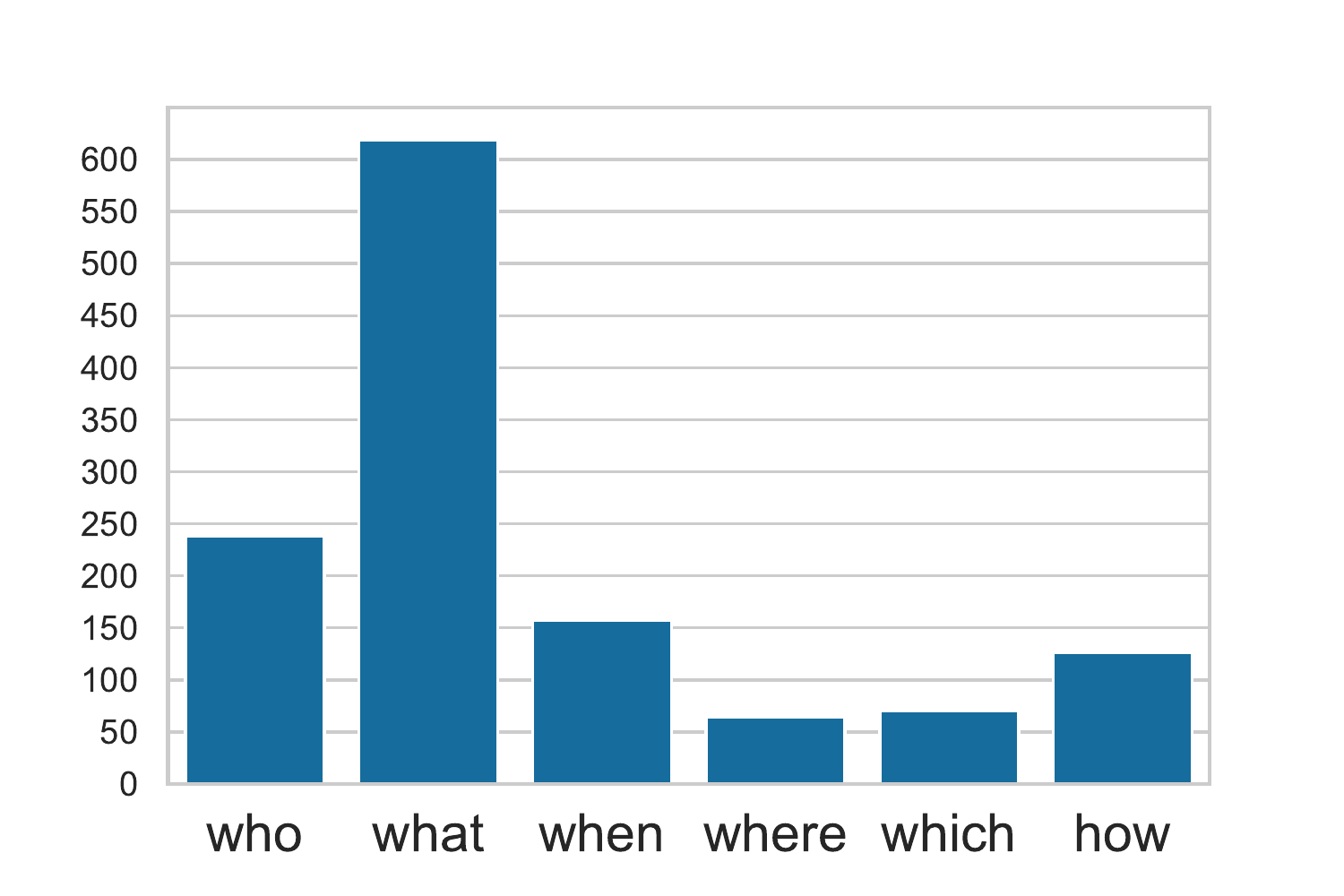}
\caption{
ELECTRA\textsubscript{random}
\label{fig:bert_random_wh}}
\end{subfigure}
\hspace{-2.0mm}
\begin{subfigure}[h]{0.1667\textwidth}
\centering
    \includegraphics[width=\textwidth]{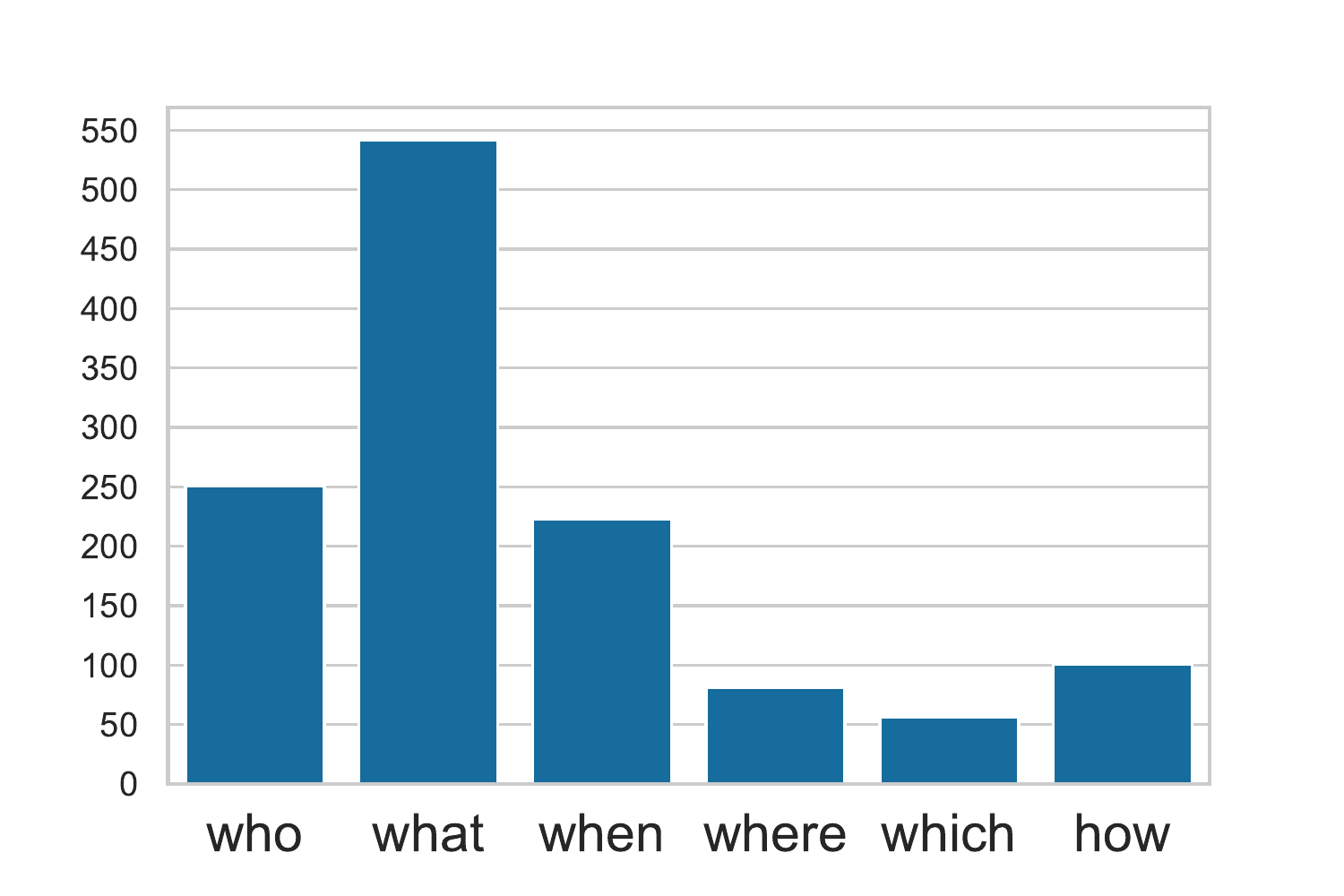}
\caption{
SDC-ELECTRA
\label{fig:bert_nomodel_wh}}
\end{subfigure}
\caption{Frequency of wh-questions generated.
\label{fig:bert_wh}}
\end{figure*}

\begin{figure*}[!t]
\centering
\begin{subfigure}[h]{0.24\textwidth}
\centering
    \includegraphics[width=\textwidth]{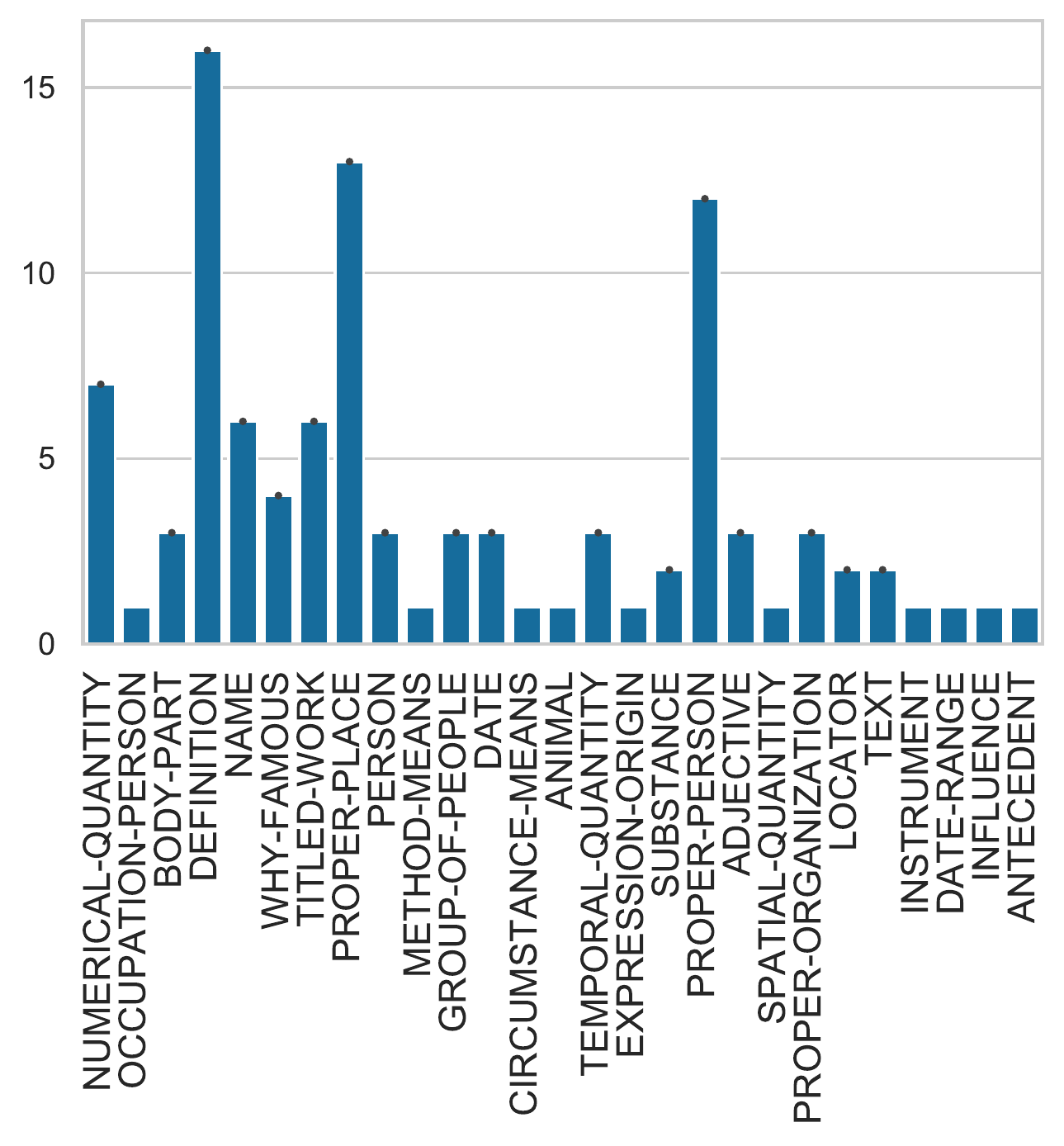}
\caption{
BERT\textsubscript{fooled}
\label{fig:bert_fooled_wh}
}
\end{subfigure}
\begin{subfigure}[h]{0.24\textwidth}
\centering
    \includegraphics[width=\textwidth]{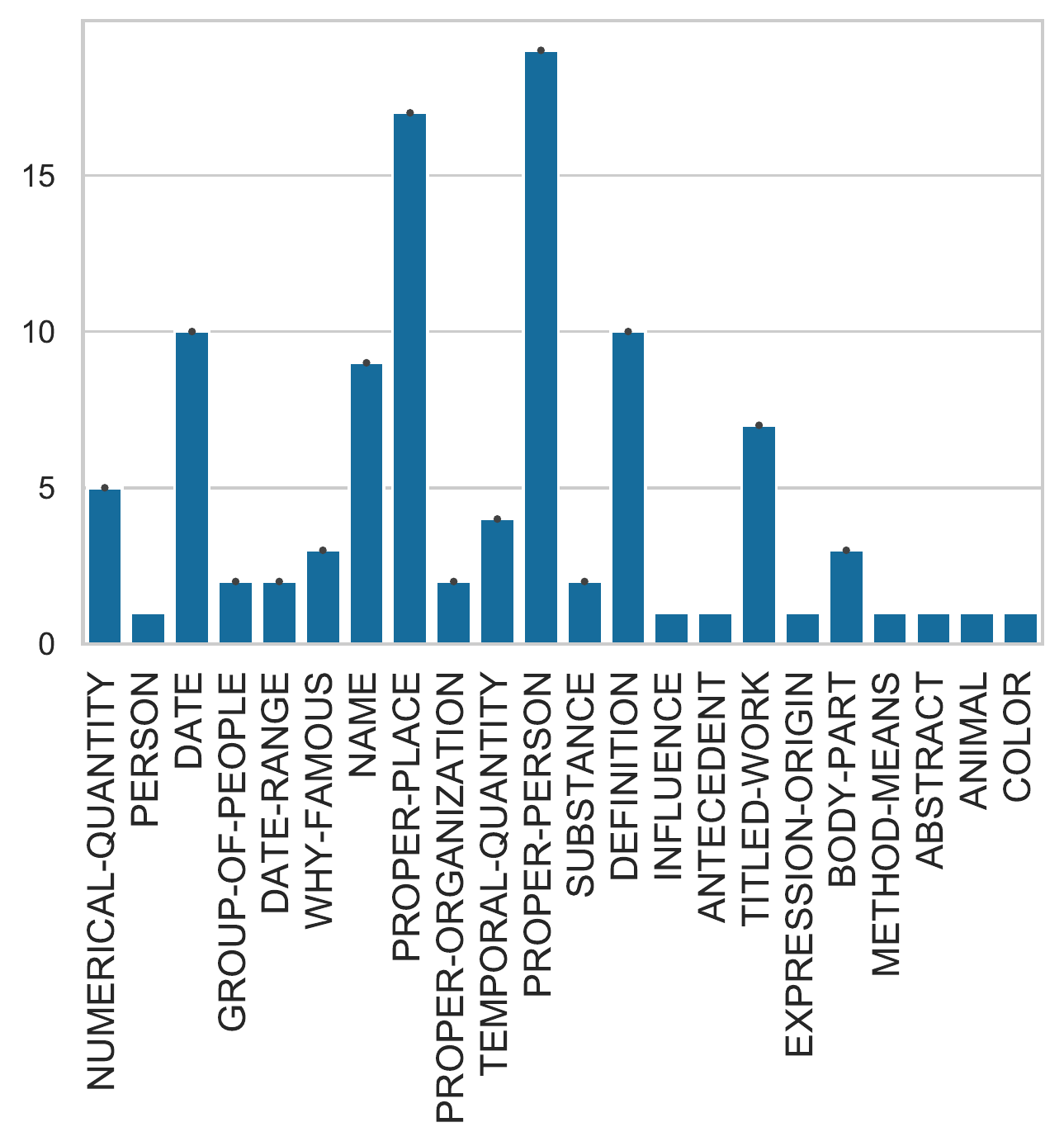}
\caption{
BERT\textsubscript{random}
\label{fig:bert_random_wh}}
\end{subfigure}
\begin{subfigure}[h]{0.24\textwidth}
\centering
    \includegraphics[width=\textwidth]{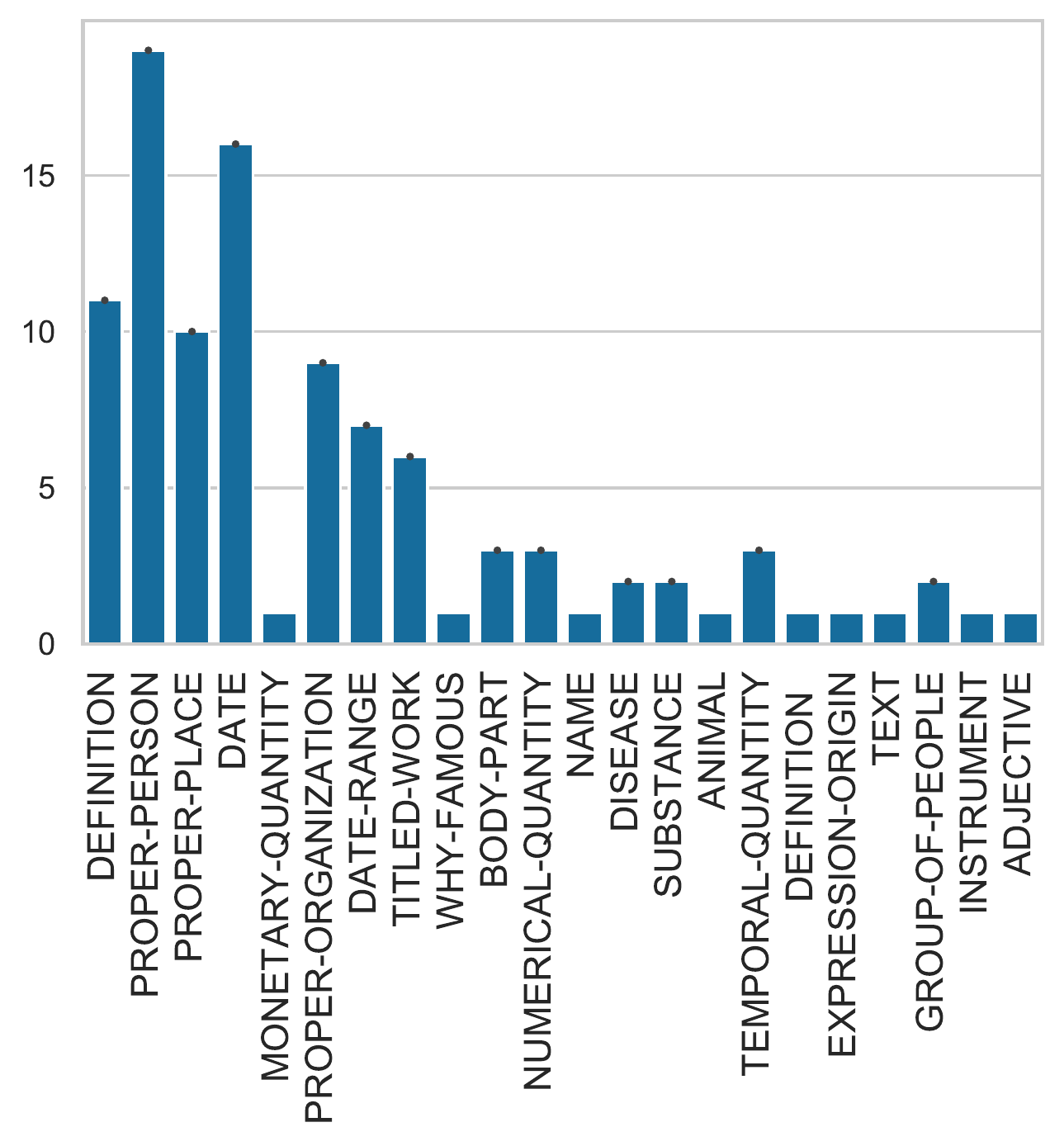}
\caption{
SDC-BERT
\label{fig:bert_nomodel_wh}}
\end{subfigure}\\
\begin{subfigure}[h]{0.24\textwidth}
\centering
    \includegraphics[width=\textwidth]{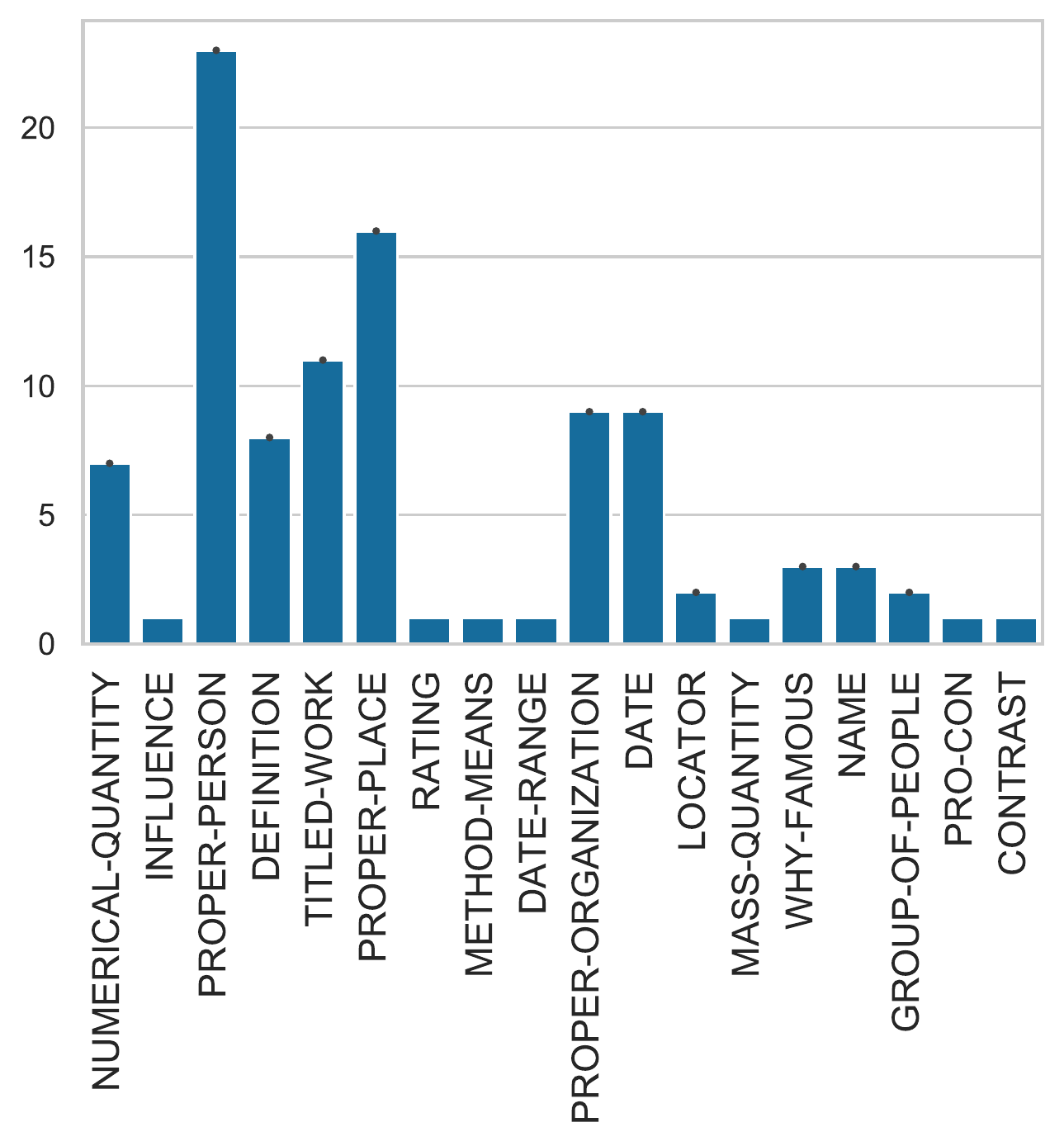}
\caption{
ELECTRA\textsubscript{fooled}
\label{fig:electra_fooled_wh}
}
\end{subfigure}
\begin{subfigure}[h]{0.24\textwidth}
\centering
    \includegraphics[width=\textwidth]{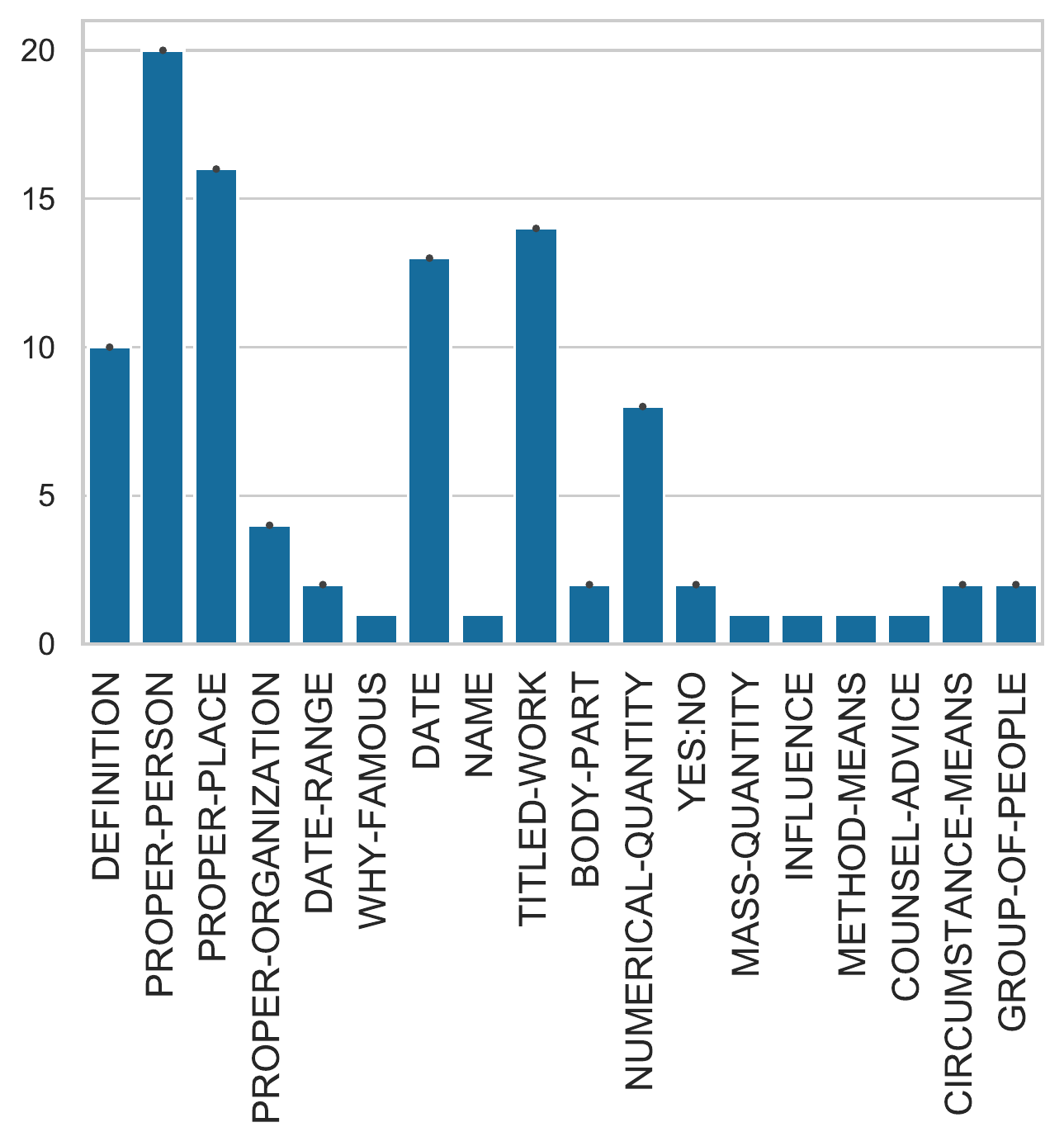}
\caption{
ELECTRA\textsubscript{random}
\label{fig:electra_random_wh}}
\end{subfigure}
\begin{subfigure}[h]{0.24\textwidth}
\centering
    \includegraphics[width=\textwidth]{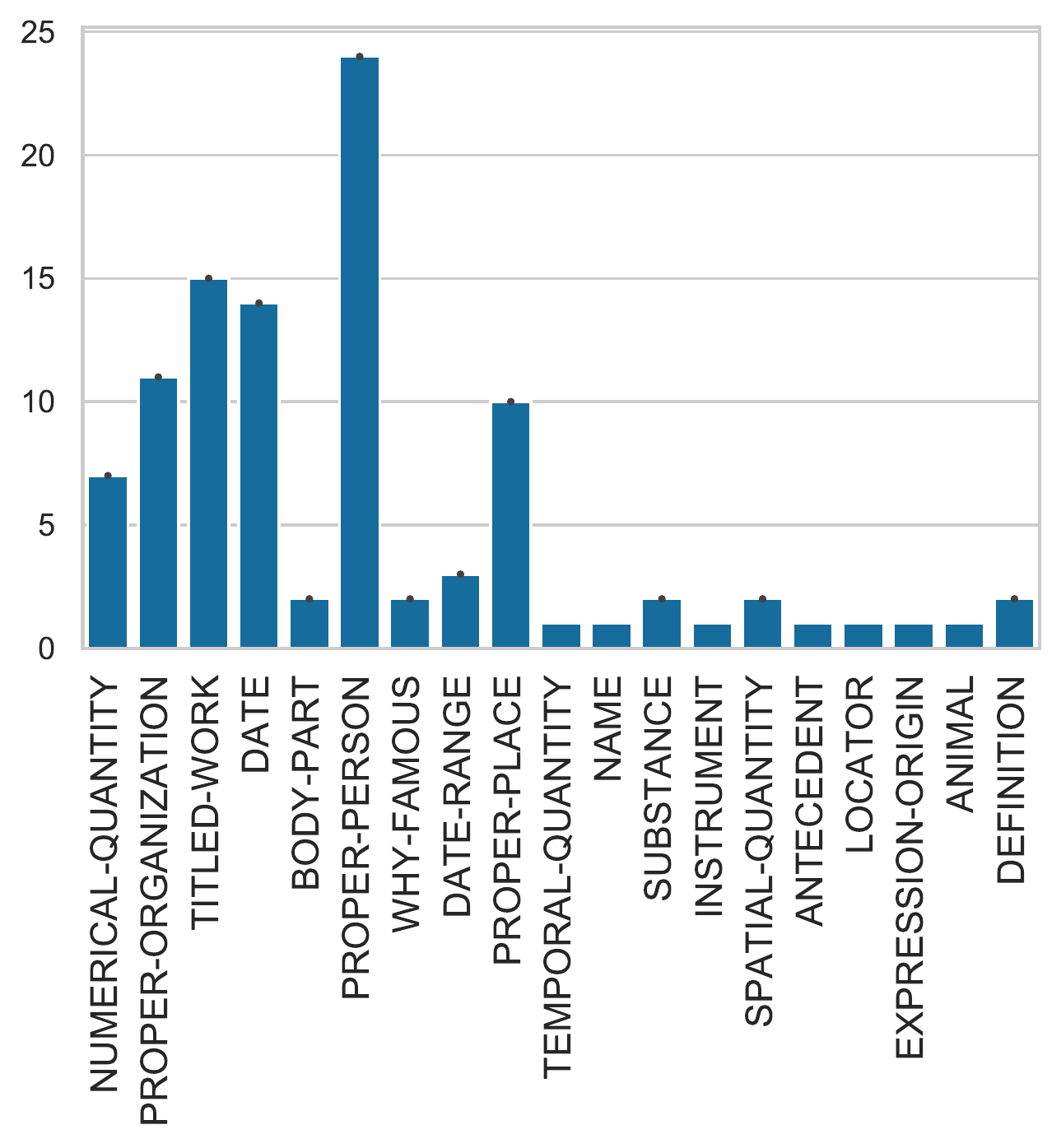}
\caption{
SDC-ELECTRA
\label{fig:electra_nomodel_wh}}
\end{subfigure}
\caption{Frequency of question types based on the taxonomy introduced by \citet{hovy2000question}.
\label{fig:qualitative_questions}}
\end{figure*}

Finally, we perform a qualitative analysis over the collected data,
revealing profound differences 
with models in (versus out of) the loop.
Recall that because these datasets 
were constructed in a randomized study, 
any observed differences are attributable 
to the model-in-the loop collection scheme.

To begin, we analyze $100$ questions from each dataset 
and categorize them using the taxonomy 
introduced by \citet{hovy2000question}.\footnote{This taxonomy can be accessed at \href{https://www.isi.edu/natural-language/projects/webclopedia/Taxonomy/taxonomy_toplevel.html}{https://www.isi.edu/nat\\ural-language/projects/webclopedia/Taxonomy/taxonomy\\\_toplevel.html}}
We also look at the first word 
of the \emph{wh}-type questions 
in each dev set 
(Fig.~\ref{fig:qualitative_questions}) 
and observe key qualitative differences 
between data via ADC and SDC for both models.

In case of ADC with BERT (and associated SDC), 
while we observe that most questions in the dev sets 
start with \emph{what}, 
ADC has a higher proportion 
compared to SDC 
($587$ in BERT\textsubscript{fooled} 
and $492$ in BERT\textsubscript{random} versus $416$ in SDC).
Furthermore, we notice that 
compared to BERT\textsubscript{fooled} dev set, 
SDC has more \emph{when}- ($148$) 
and \emph{who}-type ($220$) questions, 
the answers to which typically refer to dates, 
places and people (or organizations), respectively.
This is also reflected in the taxonomy categorization.
Interestingly, the BERT\textsubscript{random} dev set
has more \emph{when}- and \emph{who}-type questions 
than BERT\textsubscript{fooled}
($103$ and $182$ versus $50$ and $159$, respectively).
This indicates that the BERT model 
could have been better at answering questions
related to dates and people (or organizations), 
which could have further incentivized workers 
not to generate such questions 
upon observing these patterns.
Similarly, in the $100$-question samples,
we find that a larger proportion 
of questions in ADC
are categorized as requiring numerical reasoning 
($11$ and $18$ in BERT\textsubscript{fooled} 
and BERT\textsubscript{random}, respectively) 
compared to SDC ($7$).
It is possible that the model's 
performance on numerical reasoning 
(as also demonstrated by its lower performance 
on DROP compared to fine-tuning on ADC or SDC)
would have incentivized workers 
to generate more questions 
requiring numerical reasoning and as a result, 
skewed the distribution towards such questions.

Similarly, with ELECTRA, we observe 
that \emph{what}-type questions 
constitute most of the questions 
in the development sets for both ADC and SDC,
although data collected via ADC 
has a higher proportion of these
($641$ in ELECTRA\textsubscript{fooled} 
and $619$ in ELECTRA\textsubscript{random} versus $542$ in SDC).
We also notice more \emph{how}-type questions 
in ADC ($126$ in ELECTRA\textsubscript{random}) 
vs $101$ in SDC, 
and that the SDC sample has more questions 
that relate to dates ($223$) 
but the number is lower in the ADC samples 
($157$ and $86$ in ELECTRA\textsubscript{random} 
and ELECTRA\textsubscript{fooled}, respectively).
As with BERT, the ELECTRA model 
was likely better at identifying answers 
about dates or years
which could have further incentivized workers
to generate less questions of such types.
However, unlike with BERT,
we observe that the ELECTRA ADC and SDC
$100$-question samples 
contain similar numbers of questions 
involving numerical answers 
($8$, $9$ and $10$ in ELECTRA\textsubscript{fooled}, ELECTRA\textsubscript{random} and SDC respectively).

Lastly, despite explicit instructions 
not to generate questions about passage structure (Fig.~\ref{fig:annotation_platform2}), 
a small number of workers
nevertheless created such questions.
For instance, one worker wrote, 
``\emph{What is the number in the passage 
that is one digit less than 
the largest number in the passage?}''
While most such questions 
were discarded during validation,
some of these are present in the final data. 
Overall, we notice considerable differences 
between ADC and SDC data, 
particularly vis-a-vis 
what kind of questions workers generate.
Our qualitative analysis 
offers additional insights
that suggest that ADC 
would skew the distribution 
of questions workers create,
as the incentives align 
with quickly creating more questions
that can fool the model.
This is reflected in all our ADC datasets.
One remedy could be 
to provide workers with initial questions,
asking them to edit those questions
to elicit incorrect model predictions.
Similar strategies were employed in \cite{ettinger2017towards},
where \emph{breakers} minimally edited original data
to elicit incorrect predictions 
from the models built by \emph{builders},
as well as in recently introduced 
adversarial benchmarks 
for sentiment analysis \citep{potts2020dynasent}.

\section{Conclusion}
In this paper, we demonstrated 
that across a variety 
of models and datasets, 
training on adversarial data 
leads to better performance on 
evaluation sets created in a similar fashion,
but tends to yield worse performance
on out-of-domain evaluation sets
not created adversarially. 
Additionally, our results
suggest that the ADC process 
(regardless of the outcome) 
might matter
more than successfully fooling a model.
We also identify key qualitative differences
between data generated via ADC and SDC,
particularly the kinds of questions created.

Overall, our work investigates 
ADC in a controlled setting,
offering insights 
that can guide future research 
in this direction.
These findings are particularly important 
given that ADC is more time-consuming and expensive than SDC, 
with workers requiring additional financial incentives.
We believe that a remedy to these issues
could be to ask workers to edit questions
rather than to generate them.
In the future, we would like to extend this study 
and investigate the efficacy of 
various constraints on question creation, 
and the role of other factors 
such as domain complexity, passage length,
and incentive structure, among others.

\section*{Acknowledgements}
The authors thank Max Bartolo, Robin Jia, Tanya Marwah, Sanket Vaibhav Mehta, Sina Fazelpour, Kundan Krishna, Shantanu Gupta, Simran Kaur, and Aishwarya Kamath for their valuable feedback on the crowdsourcing platform and the paper.

\section*{Ethical Considerations}
The passages in our datasets are sourced from the datasets released by \citet{karpukhin2020dense} under a Creative Commons License. As described in main text, we designed our incentive structure to ensure that crowdworkers were paid $\$15$/hour, which is twice the US federal minimum wage. Our datasets focus on the English language, and are not collected for the purpose of designing NLP applications but to conduct a human study. We share our dataset to allow the community to replicate our findings and do not foresee any risks associated with the use of this data.

\bibliography{custom-norm}
\bibliographystyle{acl_natbib}

\appendix

\section{Appendix}
\label{sec:appendix}
\begingroup
\begin{table*}[t]
  \centering
  \tiny
  \begin{tabular}{ l c c c c c c c c}
    \toprule
    Evaluation set $\rightarrow$ & \multicolumn{2}{c}{ELECTRA\textsubscript{fooled}} & \multicolumn{2}{c}{ELECTRA\textsubscript{random}} & \multicolumn{2}{c}{SDC} & \multicolumn{2}{c}{Original Dev.} \\
    Training set $\downarrow$ & EM & F1 & EM & F1 & EM & F1 & EM & F1\\
     \midrule
        \multicolumn{9}{c}{Finetuned model: BERT\textsubscript{large}}\\
    \midrule
    Original (O; 23.1k) & $23.3$ & $31.9$ & $56.7$ & $72.6$ & $63.8$ & $78.5$ & $73.3$ & $80.5$\\
    Original (14.6k) & $36.7_{0.4}$ & $50.7_{0.3}$ & $48.2_{0.4}$ & $64.4_{0.2}$ & $55.7_{0.1}$ & $70.5_{0.3}$ & $67.1_{0.2}$ & $75.2_{0.1}$ \\
    \midrule
    ELECTRA\textsubscript{fooled} (F; 14.6k) & $\mathbf{25.1_{1.0}}$ & $\mathbf{42.4_{1.0}}$ & $35.4_{1.5}$ & $54.3_{1.1}$ & $39.1_{2.4}$ & $59.3_{1.7}$ & $31.9_{7.9}$ & $45.0_{9.2}$\\
    ELECTRA\textsubscript{random} (R; 14.6k) & $25.4_{1.1}$ & $42.0_{1.0}$ & $\mathbf{38.4_{0.9}}$ & $56.8_{0.8}$ & $42.0_{1.4}$ & $61.7_{1.3}$ & $46.4_{3.1}$ & $60.6_{3.8}$\\
    SDC (14.6k) & $23.1_{1.0}$ & $40.8_{1.3}$ & $36.3_{1.3}$ & $56.3_{1.3}$ & $\mathbf{45.2_{1.8}}$ & $\mathbf{65.4_{1.5}}$ & $48.6_{1.6}$ & $62.3_{1.9}$\\
    \midrule
    O + F (37.7k) & $\mathbf{26.7_{1.7}}$ & $\mathbf{43.1_{0.9}}$ & $40.1_{1.3}$ & $58.7_{1.5}$ & $44.6_{0.9}$ & $64.2_{1.2}$ & $72.1_{0.5}$ & $79.7_{0.7}$\\
    O + R  (37.7k) & $26.0_{0.8}$ & $42.9_{0.6}$ & $41.7_{0.5}$ & $60.3_{0.6}$ & $47.1_{1.4}$ & $66.5_{1.3}$ & $\mathbf{73.0_{0.5}}$ & $\mathbf{80.5_{0.2}}$\\
    O + SDC  (37.7k) & $24.5_{0.7}$ & $41.7_{0.7}$ & $41.4_{0.9}$ & $60.7_{0.4}$ & $\mathbf{50.9_{1.0}}$ & $\mathbf{69.7_{0.3}}$ & $72.0_{0.1}$ & $79.7_{0.1}$\\
    \midrule
    \multicolumn{9}{c}{Finetuned model: RoBERTa\textsubscript{large}}\\
    \midrule
    Original (O; 23.1k) & $49.2$ & $64.4$ & $59.1$ & $75.8$ & $64.5$ & $79.8$ & $73.5$ & $80.5$\\
    Original (14.6k) & $48.3_{0.9}$ & $63.3_{1.4}$ & $58.7_{0.9}$ & $74.9_{1.0}$ & $62.7_{0.4}$ & $79.0_{0.7}$ & $71.5_{0.5}$ & $79.3_{0.6}$ \\
    \midrule
    ELECTRA\textsubscript{fooled} (F; 14.6k) & $\mathbf{65.3_{0.5}}$ & $\mathbf{79.9_{0.5}}$ & $69.4_{0.6}$ & $84.6_{0.5}$ & $75.8_{0.6}$ & $89.0_{0.3}$ & $55.9_{1.2}$ & $67.5_{1.0}$\\
    ELECTRA\textsubscript{random} (R; 14.6k) & $64.6_{0.5}$ & $79.4_{0.4}$ & $\mathbf{70.4_{0.5}}$ & $\mathbf{85.4_{0.3}}$ & $76.5_{0.5}$ & $89.4_{0.3}$ & $\mathbf{59.8_{1.2}}$ & $\mathbf{70.6_{0.9}}$\\
    SDC (14.6k) & $61.0_{0.2}$ & $77.1_{0.3}$ & $67.9_{0.4}$ & $84.1_{0.4}$ & $\mathbf{77.3_{0.5}}$ & $\mathbf{89.9_{0.3}}$ & $55.7_{1.0}$ & $68.8_{0.8}$\\
    \midrule
    O + F (37.7k) & $\mathbf{65.0_{0.3}}$ & $\mathbf{79.9_{0.3}}$ & $70.1_{0.5}$ & $85.2_{0.4}$ & $76.2_{0.3}$ & $89.7_{0.2}$ & $73.3_{0.3}$ & $80.7_{0.2}$\\
    O + R  (37.7k) & $64.3_{0.3}$ & $78.8_{0.3}$ & $\mathbf{70.7_{0.2}}$ & $\mathbf{85.8_{0.2}}$ & $76.5_{0.6}$ & $89.7_{0.3}$ & $73.4_{0.5}$ & $80.8_{0.3}$\\
    O + SDC  (37.7k) & $61.5_{0.5}$ & $77.2_{0.3}$ & $69.0_{0.4}$ & $84.7_{0.4}$ & $\mathbf{77.6_{0.4}}$ & $\mathbf{90.5_{0.2}}$ & $73.6_{0.5}$ & $80.9_{0.4}$\\
    \midrule
    \multicolumn{9}{c}{Finetuned model: ELECTRA\textsubscript{large}}\\
    \midrule
    Original (O; 23.1k) & $0$ & $10.8$ & $40.2$ & $57.8$ & $44.8$ & $60.9$ & $74.2$ & $81.2$\\
    Original (14.6k) & $25.9_{0.2}$ & $40.9_{0.4}$ & $37.3_{0.6}$ & $63.9_{0.7}$ & $53.6_{1.3}$ & $74.7_{1.1}$ & $71.9_{0.3}$ & $79.5_{0.3}$ \\
    \midrule
    ELECTRA\textsubscript{fooled} (F; 14.6k) & $26.4_{1.5}$ & $44.0_{1.6}$ & $41.2_{1.5}$ & $60.8_{1.3}$ & $42.7_{4.0}$ & $63.5_{3.2}$ & $57.5_{0.9}$ & $68.8_{0.7}$ \\
    ELECTRA\textsubscript{random} (R; 14.6k) & $23.4_{4.9}$ & $40.5_{5.6}$ & $42.3_{6.9}$ & $62.3_{7.0}$ & $42.1_{8.0}$ & $62.9_{7.5}$ & $57.6_{0.8}$ & $69.3_{1.0}$ \\
    SDC (14.6k) & $24.5_{2.4}$ & $43.7_{3.5}$ & $40.6_{3.5}$ & $61.5_{3.8}$ & $46.9_{5.4}$ & $68.2_{4.7}$ & $54.9_{1.8}$ & $68.3_{1.2}$\\
    \midrule
    O + F (37.7k) & $25.3_{1.9}$ & $43.7_{2.0}$ & $40.2_{1.9}$ & $60.6_{1.9}$ & $41.7_{3.9}$ & $63.4_{3.6}$ & $73.6_{0.5}$ & $81.1_{0.4}$\\
    O + R  (37.7k) & $21.7_{1.1}$ & $40.1_{1.1}$ & $42.2_{2.3}$ & $64.8_{1.9}$ & $38.0_{3.6}$ & $60.8_{2.9}$ & $74.4_{0.3}$ & $\mathbf{81.7_{0.1}}$\\
    O + SDC  (37.7k) & $24.5_{1.8}$ & $43.4_{1.6}$ & $42.8_{1.5}$ & $63.5_{1.0}$ & $\mathbf{49.6_{1.9}}$ & $\mathbf{70.3_{1.5}}$ & $74.2_{0.2}$ & $81.5_{0.1}$\\
\bottomrule
  \end{tabular}
  \caption{EM and F1 scores of various models evaluated on adversarial datasets collected with an ELECTRA\textsubscript{large} model and non-adversarial datasets. Adversarial results in bold are statistically significant compared to SDC setting and vice versa with $p<0.05.$
  \label{tab:ts_collected_electra}}
\end{table*}
\endgroup

\begin{table*}[th]
  \centering
  \tiny
  \begin{tabular}{ l c c c c c c}
    \toprule
    Evaluation set $\rightarrow$ & \multicolumn{2}{c}{D\textsubscript{RoBERTa}} & \multicolumn{2}{c}{D\textsubscript{BERT}} & \multicolumn{2}{c}{D\textsubscript{BiDAF}} \\
    Training set $\downarrow$ & EM & F1 & EM & F1 & EM & F1\\
     \midrule
        \multicolumn{7}{c}{Finetuned model: BERT\textsubscript{large}}\\
    \midrule
    Original (23.1k) & $6.0$ & $13.5$ & $8.1$ & $14.2$ & $12.6$ & $21.4$\\
    Original (14.6k) & $5.3_{0.2}$ & $11.4_{0.2}$ & $6.8_{0.8}$ & $13.9_{0.5}$ & $12.1_{0.4}$ & $20.6_{0.2}$ \\
    
    \midrule
    
    ELECTRA\textsubscript{fooled}14.6k) & $3.8_{0.5}$ & $13.3_{0.7}$ & $6.2_{0.7}$ & $16.4_{0.5}$ & $12.6_{1.2}$ & $26.2_{1.0}$\\
    ELECTRA\textsubscript{random}14.6k) & $\mathbf{4.3_{0.5}}$ & $13.7_{0.7}$ & $\mathbf{6.4_{0.4}}$ & $\mathbf{16.4_{0.8}}$ & $\mathbf{13.6_{0.8}}$ & $\mathbf{27.1_{1.2}}$\\
    SDC (14.6k) & $3.9_{0.4}$ & $13.2_{0.4}$ & $5.4_{0.4}$ & $15.1_{0.5}$ & $10.8_{0.7}$ & $23.8_{0.8}$\\
    
    \midrule
    
    Orig + ELECTRA\textsubscript{fooled} (37.7k) & $6.4_{0.5}$ & $16.1_{0.3}$ & $7.8_{0.8}$ & $18.0_{0.6}$ & $17.0_{0.2}$ & $31.0_{0.6}$\\
    Orig + ELECTRA\textsubscript{random}  (37.7k) & $\mathbf{6.6_{0.6}}$ & $\mathbf{16.1_{0.3}}$ & $8.5_{0.6}$ & $18.4_{0.5}$ & $16.9_{0.3}$ & $30.8_{0.4}$\\
    Orig + SDC  (37.7k) & $5.8_{0.2}$ & $15.6_{0.4}$ & $8.7_{0.5}$ & $18.7_{0.6}$ & $17.4_{0.7}$ & $30.0_{0.8}$\\
    
    \midrule
    
    \multicolumn{7}{c}{Finetuned model: RoBERTa\textsubscript{large}}\\
    \midrule
    Original (23.1k) & $15.7$ & $25.0$ & $26.5$ & $37.0$ & $37.9$ & $50.4$\\
    Original (14.6k) & $14.3_{0.2}$ & $23.7_{0.3}$ & $25.1_{0.3}$ & $35.4_{0.7}$ & $37.4_{0.7}$ & $50.2_{0.5}$ \\

    \midrule
    
    ELECTRA\textsubscript{fooled}14.6k) & $\mathbf{16.4_{0.9}}$ & $\mathbf{27.7_{1.2}}$ & $27.4_{1.3}$ & $40.8_{1.5}$ & $46.8_{1.1}$ & $62.4_{1.1}$\\
    ELECTRA\textsubscript{random}14.6k) & $15.8_{1.4}$ & $27.2_{1.4}$ & $\mathbf{28.1_{1.6}}$ & $\mathbf{41.5_{1.8}}$ & $\mathbf{48.0_{0.9}}$ & $\mathbf{63.0_{0.6}}$\\
    SDC (14.6k) & $12.1_{1.0}$ & $23.9_{1.3}$ & $22.7_{1.1}$ & $35.4_{1.5}$ & $40.5_{1.3}$ & $56.8_{1.3}$\\
    
    \midrule
    
    Orig + ELECTRA\textsubscript{fooled} (37.7k) & $18.9_{0.8}$ & $30.4_{0.9}$ & $\mathbf{33.2_{0.8}}$ & $\mathbf{46.4_{0.6}}$ & $\mathbf{49.2_{0.9}}$ & $\mathbf{65.1_{0.8}}$\\
    Orig + ELECTRA\textsubscript{random}  (37.7k) & $18.0_{0.4}$ & $29.6_{0.3}$ & $32.3_{0.6}$ & $45.1_{1.2}$ & $48.2_{0.8}$ & $63.5_{0.6}$\\
    Orig + SDC  (37.7k) & $18.2_{1.0}$ & $29.7_{0.9}$ & $28.2_{0.3}$ & $41.4_{0.5}$ & $45.0_{0.9}$ & $60.9_{0.6}$\\
    
    \midrule
    
    \multicolumn{7}{c}{Finetuned model: ELECTRA\textsubscript{large}}\\
    \midrule
    Original (23.1k) & $8.2$ & $17.4$ & $15.7$ & $24.2$ & $22.4$ & $34.3$\\
    Original (14.6k) & $9.5_{0.2}$ & $18.0_{0.5}$ & $15.4_{0.5}$ & $24.2_{0.6}$ & $21.7_{0.2}$ & $33.1_{0.1}$ \\

    \midrule
    
    ELECTRA\textsubscript{fooled}14.6k) & $10.2_{0.3}$ & $21.7_{0.5}$ & $17.0_{0.7}$ & $29.7_{0.6}$ & $21.7_{1.7}$ & $36.6_{1.1}$\\
    ELECTRA\textsubscript{random}14.6k) & $10.4_{0.5}$ & $21.3_{0.5}$ & $16.5_{0.2}$ & $28.6_{0.8}$ & $19.9_{5.0}$ & $34.4_{5.9}$\\
    SDC (14.6k) & $10.3_{0.8}$ & $21.6_{0.7}$ & $15.8_{1.1}$ & $28.5_{1.2}$ & $19.3_{4.8}$ & $33.3_{7.8}$\\
    
    \midrule
    
    Orig + ELECTRA\textsubscript{fooled} (37.7k) & $10.2_{0.3}$ & $21.7_{0.5}$ & $17.0_{0.7}$ & $29.7_{0.6}$ & $24.0_{0.7}$ & $39.2_{0.7}$\\
    Orig + ELECTRA\textsubscript{random}  (37.7k) & $10.4_{0.5}$ & $21.3_{0.5}$ & $16.5_{0.2}$ & $28.6_{0.8}$ & $23.5_{0.5}$ & $38.4_{0.4}$\\
    Orig + SDC  (37.7k) & $10.3_{0.8}$ & $21.6_{0.7}$ & $15.8_{1.1}$ & $28.5_{1.2}$ & $\mathbf{24.5_{0.6}}$ & $\mathbf{39.9_{0.6}}$\\
\bottomrule
  \end{tabular}
  \caption{EM and F1 scores of various models evaluated on dev datasets of \citet{bartolo2020beat}. Adversarial results in bold are statistically significant compared to SDC setting and vice versa with $p<0.05.$
  \label{tab:ts_beattheai_electra}}
\end{table*}

\begingroup
\setlength{\tabcolsep}{3.35pt}

\begin{table*}[!ht]
  \centering
  \tiny
  \begin{tabular}{ l c c c c c c c c c c c c}
    \toprule
    \multicolumn{13}{c}{Finetuned model: BERT\textsubscript{large}}\\
    \midrule
    Evaluation set $\rightarrow$ & \multicolumn{2}{c}{BioASQ} & \multicolumn{2}{c}{DROP} & \multicolumn{2}{c}{DuoRC} & \multicolumn{2}{c}{Relation Extraction} & \multicolumn{2}{c}{RACE} & \multicolumn{2}{c}{TextbookQA}\\
    Training set $\downarrow$ & EM & F1 & EM & F1 & EM & F1 & EM & F1 & EM & F1 & EM & F1\\
     \midrule
    Original (23.1k) & $19.4$ & $32.5$ & $7.8$ & $16.2$ & $14.5$ & $22.8$ & $32.0$ & $47.1$ & $11.4$ & $18.8$ & $25.0$ & $33.4$\\
    Original (14.6k) & $20.4_{0.3}$ & $35.9_{0.7}$ & $5.1_{0.3}$ & $12.4_{0.3}$ & $11.6_{0.4}$ & $17.8_{0.6}$ & $33.0_{0.9}$ & $44.2_{2.0}$ & $10.4_{0.6}$ & $17.7_{0.9}$ & $19.5_{0.6}$ & $27.3_{0.7}$ \\
     \midrule
    ELECTRA\textsubscript{fooled} (14.6k) & $13.6_{0.9}$ & $29.1_{1.1}$ & $3.2_{0.4}$ & $11.9_{0.7}$ & $11.0_{0.9}$ & $19.3_{0.6}$ & $33.6_{2.2}$ & $52.5_{2.3}$ & $7.9_{0.7}$ & $17.7_{0.8}$ & $12.2_{1.7}$ & $21.2_{1.8}$\\
    ELECTRA\textsubscript{random} (14.6k) & $15.9_{0.8}$ & $32.0_{1.7}$ & $3.1_{0.4}$ & $10.5_{0.9}$ & $12.1_{0.9}$ & $20.4_{1.4}$ & $35.7_{3.1}$ & $55.6_{3.7}$ & $9.5_{0.7}$ & $19.1_{0.8}$ & $14.6_{1.8}$ & $23.9_{1.8}$\\
    SDC (14.6k) & $\mathbf{17.1_{0.7}}$ & $\mathbf{34.5_{1.0}}$ & $2.6_{0.3}$ & $10.1_{0.9}$ & $11.9_{0.8}$ & $21.2_{1.2}$ & $34.2_{3.4}$ & $53.7_{4.1}$ & $9.2_{1.0}$ & $19.0_{0.7}$ & $\mathbf{17.5_{1.1}}$ & $\mathbf{27.4_{1.3}}$\\
    \midrule
    Orig + Fooled (37.7k) & $17.8_{1.0}$ & $33.5_{2.0}$ & $6.1_{1.1}$ & $16.1_{1.7}$ & $14.2_{1.4}$ & $22.9_{1.9}$ & $42.0_{2.2}$ & $59.6_{2.5}$ & $12.0_{0.9}$ & $22.2_{0.9}$ & $24.6_{1.0}$ & $33.7_{1.2}$\\
    Orig + Random (37.7k) & $20.0_{1.1}$ & $36.4_{1.6}$ & $6.8_{0.9}$ & $17.1_{1.0}$ & $14.6_{1.0}$ & $23.5_{1.5}$ & $44.0_{1.3}$ & $61.8_{1.3}$ & $12.0_{0.9}$ & $22.0_{0.9}$ & $23.9_{0.8}$ & $33.5_{1.0}$\\
    Orig + SDC (37.7k) & $\mathbf{21.8_{0.6}}$ & $\mathbf{39.2_{1.1}}$ & $6.1_{0.5}$ & $16.1_{0.7}$ & $\mathbf{16.7_{0.9}}$ & $\mathbf{25.9_{1.0}}$ & $\mathbf{43.4_{0.7}}$ & $61.0_{1.1}$ & $11.9_{0.7}$ & $22.5_{0.7}$ & $\mathbf{25.4_{0.5}}$ & $\mathbf{35.5_{0.6}}$\\
    \midrule
    & \multicolumn{2}{c}{HotpotQA}  & \multicolumn{2}{c}{Natural Questions} & \multicolumn{2}{c}{NewsQA} & \multicolumn{2}{c}{SearchQA} & \multicolumn{2}{c}{SQuAD} & \multicolumn{2}{c}{TriviaQA}\\
    & EM & F1 & EM & F1 & EM & F1 & EM & F1 & EM & F1 & EM & F1\\
     \midrule
    Original (23.1k) & $19.4$ & $33.9$ & $36.3$ & $48.7$ & $16.2$ & $25.6$ & $11.3$ & $19.3$ & $32.5$ & $46.0$ & $16.8$ & $25.3$ \\
    Original (14.6k) & $17.4_{0.9}$ & $28.7_{1.2}$ & $35.0_{0.7}$ & $47.7_{0.7}$ & $12.8_{0.2}$ & $22.6_{0.1}$ & $9.0_{0.1}$ & $13.8_{0.4}$ & $26.0_{0.3}$ & $39.2_{0.7}$ & $11.8_{0.5}$ & $18.2_{0.7}$ \\
     \midrule
    ELECTRA\textsubscript{fooled} (14.6k) & $19.1_{0.7}$ & $33.4_{0.8}$ & $28.0_{1.4}$ & $43.1_{1.4}$ & $12.9_{0.8}$ & $25.9_{0.8}$ & $4.0_{0.3}$ & $9.1_{0.5}$ & $26.9_{1.4}$ & $46.4_{1.4}$ & $9.2_{0.8}$ & $16.3_{1.1}$\\
    ELECTRA\textsubscript{random} (14.6k) & $21.2_{1.0}$ & $35.5_{1.3}$ & $29.0_{2.3}$ & $43.8_{2.3}$ & $13.8_{0.8}$ & $27.1_{1.3}$ & $4.2_{0.4}$ & $9.1_{0.6}$ & $29.2_{1.6}$ & $48.3_{2.2}$ & $10.0_{0.7}$ & $17.3_{1.2}$\\
    SDC (14.6k) & $\mathbf{23.5_{1.2}}$ & $\mathbf{37.8_{1.3}}$ & $28.4_{1.7}$ & $43.5_{1.4}$ & $\mathbf{15.6_{0.8}}$ & $\mathbf{30.3_{1.0}}$ & $\mathbf{5.0_{0.5}}$ & $\mathbf{9.9_{0.7}}$ & $\mathbf{31.5_{0.7}}$ & $\mathbf{50.5_{0.8}}$ & $10.0_{0.9}$ & $\mathbf{19.1_{1.3}}$\\
    \midrule
    Orig + Fooled (37.7k) & $25.5_{1.4}$ & $40.8_{1.5}$ & $38.5_{1.1}$ & $52.2_{1.1}$ & $17.0_{0.7}$ & $30.9_{1.2}$ & $9.9_{0.4}$ & $15.8_{0.8}$ & $32.7_{1.5}$ & $51.7_{1.5}$ & $14.2_{1.6}$ & $22.6_{1.8}$\\
    Orig + Random (37.7k) & $26.7_{1.2}$ & $41.9_{1.2}$ & $38.6_{1.0}$ & $52.6_{0.7}$ & $17.0_{0.4}$ & $30.7_{0.7}$ & $9.2_{0.9}$ & $14.6_{1.2}$ & $34.3_{0.6}$ & $53.3_{0.8}$ & $14.1_{0.7}$ & $22.7_{1.1}$\\
    Orig + SDC (37.7k) & $29.0_{1.0}$ & $42.6_{0.8}$ & $38.7_{0.3}$ & $52.4_{0.1}$ & $\mathbf{18.7_{0.6}}$ & $\mathbf{33.9_{0.5}}$ & $\mathbf{11.1_{0.7}}$ & $\mathbf{16.6_{0.9}}$ & $\mathbf{36.1_{0.7}}$ & $\mathbf{54.9_{0.5}}$ & $\mathbf{15.1_{0.3}}$ & $\mathbf{24.2_{0.2}}$\\
    \midrule
    \multicolumn{13}{c}{Finetuned model: RoBERTa\textsubscript{large}}\\
    \midrule
    Evaluation set $\rightarrow$ & \multicolumn{2}{c}{BioASQ} & \multicolumn{2}{c}{DROP} & \multicolumn{2}{c}{DuoRC} & \multicolumn{2}{c}{Relation Extraction} & \multicolumn{2}{c}{RACE} & \multicolumn{2}{c}{TextbookQA}\\
    Training set $\downarrow$ & EM & F1 & EM & F1 & EM & F1 & EM & F1 & EM & F1 & EM & F1\\
     \midrule
    Original (23.1k) & $47.7$ & $63.5$ & $37.2$ & $48.1$ & $38.6$ & $49.1$ & $74.4$ & $85.9$ & $33.7$ & $44.9$ & $36.4$ & $46$ \\
    Original (14.6k) & $45.4_{1.7}$ & $61.8_{1.0}$ & $37.5_{1.7}$ & $48.7_{2.0}$ & $37.8_{0.7}$ & $48.7_{0.8}$ & $75.0_{0.6}$ & $86.0_{0.2}$ & $32.4_{0.7}$ & $43.4_{0.9}$ & $36.8_{1.1}$ & $46.2_{1.3}$ \\
     \midrule
    ELECTRA\textsubscript{fooled} (14.6k) & $41.2_{1.4}$ & $57.2_{1.1}$ & $30.3_{1.7}$ & $44.9_{1.8}$ & $37.9_{2.1}$ & $47.2_{2.3}$ & $74.1_{0.8}$ & $86.0_{0.4}$ & $31.7_{1.3}$ & $45.4_{1.0}$ & $30.8_{1.7}$ & $40.5_{1.8}$\\
    ELECTRA\textsubscript{random} (14.6k) & $43.3_{1.4}$ & $60.0_{1.5}$ & $\mathbf{34.1_{2.4}}$ & $\mathbf{48.8_{2.0}}$ & $39.2_{1.5}$ & $48.8_{1.6}$ & $75.5_{0.5}$ & $85.9_{0.2}$ & $\mathbf{32.6_{0.7}}$ & $46.3_{0.5}$ & $32.2_{1.2}$ & $42.2_{1.4}$\\
    SDC (14.6k) & $43.7_{1.0}$ & $\mathbf{62.5_{0.7}}$ & $27.5_{2.6}$ & $43.4_{2.9}$ & $\mathbf{42.3_{0.9}}$ & $\mathbf{53.5_{1.1}}$ & $74.9_{0.8}$ & $85.3_{0.7}$ & $31.5_{0.9}$ & $46.0_{1.0}$ & $\mathbf{36.3_{2.0}}$ & $\mathbf{47.2_{2.0}}$\\
    \midrule
    Orig + Fooled (37.7k) & $45.0_{1.2}$ & $61.2_{1.0}$ & $\mathbf{45.9_{1.6}}$ & $\mathbf{58.1_{1.3}}$ & $36.8_{1.4}$ & $47.2_{1.7}$ & $73.9_{0.4}$ & $86.3_{0.3}$ & $33.7_{0.9}$ & $47.3_{0.9}$ & $38.5_{0.9}$ & $48.3_{1.2}$\\
    Orig + Random (37.7k) & $46.3_{1.0}$ & $62.6_{0.8}$ & $45.5_{1.2}$ & $57.8_{0.8}$ & $39.1_{1.3}$ & $49.3_{1.3}$ & $74.7_{0.5}$ & $86.6_{0.2}$ & $34.1_{0.2}$ & $47.2_{0.4}$ & $39.9_{1.5}$ & $49.9_{1.9}$\\
    Orig + SDC (37.7k) & $\mathbf{47.5_{0.5}}$ & $\mathbf{64.0_{0.5}}$ & $42.7_{1.1}$ & $55.5_{1.0}$ & $\mathbf{42.1_{1.3}}$ & $\mathbf{53.7_{1.1}}$ & $74.7_{0.9}$ & $86.9_{0.5}$ & $33.9_{1.2}$ & $47.3_{1.0}$ & $\mathbf{41.9_{0.4}}$ & $\mathbf{52.5_{0.3}}$\\
    \midrule
    & \multicolumn{2}{c}{HotpotQA}  & \multicolumn{2}{c}{Natural Questions} & \multicolumn{2}{c}{NewsQA} & \multicolumn{2}{c}{SearchQA} & \multicolumn{2}{c}{SQuAD} & \multicolumn{2}{c}{TriviaQA}\\
    & EM & F1 & EM & F1 & EM & F1 & EM & F1 & EM & F1 & EM & F1\\
     \midrule
    Original (23.1k) & $19.4$ & $33.9$ & $36.3$ & $48.7$ & $16.2$ & $25.6$ & $11.3$ & $19.3$ & $32.5$ & $46.0$ & $16.8$ & $25.3$ \\
    Original (14.6k) & $47.0_{0.3}$ & $62.7_{0.3}$ & $55.6_{0.4}$ & $67.5_{0.5}$ & $38.2_{0.2}$ & $53.6_{0.3}$ & $34.5_{0.8}$ & $43.8_{0.6}$ & $60.5_{0.4}$ & $75.6_{0.5}$ & $46.5_{0.5}$ & $58.5_{0.7}$ \\
     \midrule
    ELECTRA\textsubscript{fooled} (14.6k) & $51.9_{0.9}$ & $67.9_{1.0}$ & $49.6_{0.6}$ & $64.1_{0.7}$ & $37.8_{0.9}$ & $54.9_{1.0}$ & $24.0_{2.0}$ & $31.3_{2.2}$ & $66.2_{0.4}$ & $82.0_{0.3}$ & $45.1_{1.1}$ & $55.2_{1.1}$\\
    ELECTRA\textsubscript{random} (14.6k) & $54.5_{0.8}$ & $71.0_{0.8}$ & $51.6_{0.6}$ & $65.9_{0.6}$ & $40.2_{1.1}$ & $57.7_{1.2}$ & $24.3_{2.6}$ & $32.9_{2.6}$ & $66.9_{0.2}$ & $82.6_{0.2}$ & $45.8_{0.8}$ & $56.2_{1.0}$\\
    SDC (14.6k) & $\mathbf{55.8_{0.8}}$ & $71.8_{0.8}$ & $51.7_{0.5}$ & $65.8_{0.5}$ & $\mathbf{43.9_{0.8}}$ & $\mathbf{62.1_{1.0}}$ & $24.4_{2.4}$ & $32.9_{2.4}$ & $\mathbf{68.4_{0.5}}$ & $\mathbf{84.3_{0.3}}$ & $\mathbf{47.3_{0.7}}$ & $\mathbf{59.1_{0.7}}$\\
    \midrule
    Orig + Fooled (37.7k) & $55.6_{0.8}$ & $71.7_{0.9}$ & $57.1_{0.3}$ & $69.6_{0.3}$ & $40.6_{1.5}$ & $57.7_{1.8}$ & $38.3_{2.4}$ & $47.3_{2.7}$ & $67.0_{0.5}$ & $82.7_{0.4}$ & $46.7_{1.0}$ & $57.5_{1.0}$\\
    Orig + Random (37.7k) & $56.0_{0.2}$ & $71.9_{0.3}$ & $56.5_{0.2}$ & $69.1_{0.3}$ & $42.3_{0.3}$ & $59.3_{0.7}$ & $39.4_{1.6}$ & $48.5_{1.7}$ & $68.0_{0.2}$ & $83.3_{0.2}$ & $47.8_{0.3}$ & $58.8_{0.3}$\\
    Orig + SDC (37.7k) & $\mathbf{57.5_{0.7}}$ & $\mathbf{72.8_{0.6}}$ & $56.9_{0.3}$ & $69.4_{0.3}$ & $\mathbf{44.3_{0.7}}$ & $\mathbf{62.7_{0.7}}$ & $39.3_{1.0}$ & $48.6_{1.1}$ & $\mathbf{69.9_{0.4}}$ & $\mathbf{84.3_{0.2}}$ & $\mathbf{48.6_{0.5}}$ & $\mathbf{60.1_{0.5}}$\\
    \midrule
    \multicolumn{13}{c}{Finetuned model: ELECTRA\textsubscript{large}}\\
    \midrule
    Evaluation set $\rightarrow$ & \multicolumn{2}{c}{BioASQ} & \multicolumn{2}{c}{DROP} & \multicolumn{2}{c}{DuoRC} & \multicolumn{2}{c}{Relation Extraction} & \multicolumn{2}{c}{RACE} & \multicolumn{2}{c}{TextbookQA}\\
    Training set $\downarrow$ & EM & F1 & EM & F1 & EM & F1 & EM & F1 & EM & F1 & EM & F1\\
     \midrule
    Original (23.1k) & $29.1$ & $42.8$ & $17.6$ & $26.9$ & $18.9$ & $27.1$ & $53.4$ & $67.4$ & $19.6$ & $28.5$ & $32.5$ & $41.8$ \\
    Original (14.6k) & $35.4_{0.4}$ & $51.0_{0.8}$ & $16.2_{0.5}$ & $26.6_{0.8}$ & $18.8_{0.4}$ & $26.7_{0.8}$ & $46.2_{1.3}$ & $61.1_{1.7}$ & $17.3_{0.9}$ & $27.9_{0.6}$ & $29.6_{0.6}$ & $37.8_{0.7}$ \\
     \midrule
    ELECTRA\textsubscript{fooled} (14.6k) & $25.3_{1.1}$ & $41.0_{1.6}$ & $7.6_{0.9}$ & $18.9_{1.4}$ & $12.3_{1.5}$ & $20.5_{2.0}$ & $42.1_{2.0}$ & $61.4_{2.3}$ & $13.5_{0.6}$ & $25.1_{1.0}$ & $20.8_{2.5}$ & $29.5_{2.9}$\\
    ELECTRA\textsubscript{random} (14.6k) & $25.5_{4.9}$ & $41.6_{5.5}$ & $7.8_{2.6}$ & $19.2_{5.3}$ & $12.1_{2.3}$ & $19.7_{2.9}$ & $40.3_{7.7}$ & $57.7_{9.4}$ & $13.0_{2.7}$ & $24.0_{3.7}$ & $20.3_{3.5}$ & $28.8_{3.4}$\\
    SDC (14.6k) & $25.0_{7.5}$ & $41.0_{1.7}$ & $5.9_{2.1}$ & $17.9_{4.4}$ & $13.2_{3.0}$ & $22.5_{4.9}$ & $42.7_{6.6}$ & $61.9_{7.5}$ & $13.4_{2.7}$ & $24.7_{4.0}$ & $20.8_{3.8}$ & $29.5_{3.4}$\\
    \midrule
    Orig + Fooled (37.7k) & $28.4_{2.0}$ & $45.2_{2.6}$ & $15.6_{0.8}$ & $28.6_{1.0}$ & $13.3_{1.0}$ & $21.2_{1.7}$ & $41.5_{2.8}$ & $60.5_{3.3}$ & $17.6_{0.7}$ & $29.6_{0.9}$ & $32.2_{0.9}$ & $41.6_{1.1}$\\
    Orig + Random (37.7k) & $28.6_{1.6}$ & $44.9_{2.0}$ & $16.3_{0.6}$ & $29.0_{1.2}$ & $12.8_{1.0}$ & $20.9_{1.6}$ & $39.4_{3.3}$ & $58.8_{3.6}$ & $16.6_{1.3}$ & $29.0_{1.1}$ & $32.4_{0.4}$ & $42.2_{0.5}$\\
    Orig + SDC (37.7k) & $29.7_{1.9}$ & $47.0_{2.2}$ & $15.6_{0.8}$ & $29.1_{1.3}$ & $\mathbf{16.4_{0.7}}$ & $\mathbf{27.1_{0.8}}$ & $\mathbf{48.0_{1.8}}$ & $\mathbf{67.0_{1.5}}$ & $\mathbf{19.0_{0.6}}$ & $\mathbf{32.1_{0.8}}$ & $\mathbf{33.7_{0.4}}$ & $\mathbf{43.8_{0.9}}$\\
    \midrule
    & \multicolumn{2}{c}{HotpotQA}  & \multicolumn{2}{c}{Natural Questions} & \multicolumn{2}{c}{NewsQA} & \multicolumn{2}{c}{SearchQA} & \multicolumn{2}{c}{SQuAD} & \multicolumn{2}{c}{TriviaQA}\\
    & EM & F1 & EM & F1 & EM & F1 & EM & F1 & EM & F1 & EM & F1\\
     \midrule
    Original (23.1k) & $19.4$ & $33.9$ & $36.3$ & $48.7$ & $16.2$ & $25.6$ & $11.3$ & $19.3$ & $32.5$ & $46.0$ & $16.8$ & $25.3$ \\
    Original (14.6k) & $23.2_{1.0}$ & $40.2_{1.1}$ & $33.4_{0.8}$ & $49.8_{0.5}$ & $17.9_{0.5}$ & $31.1_{0.9}$ & $16.0_{0.5}$ & $22.3_{1.1}$ & $31.1_{0.4}$ & $50.1_{0.5}$ & $21.0_{0.9}$ & $29.8_{1.3}$ \\
     \midrule
    ELECTRA\textsubscript{fooled} (14.6k) & $26.2_{0.9}$ & $42.2_{0.9}$ & $31.5_{1.4}$ & $49.7_{1.1}$ & $18.7_{1.2}$ & $32.1_{1.6}$ & $6.5_{0.7}$ & $10.4_{1.0}$ & $34.5_{1.3}$ & $53.7_{1.5}$ & $13.2_{1.0}$ & $21.5_{1.3}$\\
    ELECTRA\textsubscript{random} (14.6k) & $24.7_{5.5}$ & $40.9_{6.9}$ & $27.9_{6.8}$ & $45.7_{7.6}$ & $17.2_{3.1}$ & $30.8_{3.8}$ & $6.4_{1.6}$ & $10.3_{2.1}$ & $34.1_{5.8}$ & $53.1_{6.2}$ & $12.4_{3.4}$ & $20.1_{4.5}$\\
    SDC (14.6k) & $24.4_{3.3}$ & $41.7_{5.2}$ & $28.8_{6.2}$ & $46.7_{8.3}$ & $19.2_{3.6}$ & $\mathbf{35.5_{3.2}}$ & $\mathbf{8.3_{0.9}}$ & $\mathbf{12.8_{1.6}}$ & $34.7_{4.2}$ & $54.1_{5.1}$ & $13.4_{2.0}$ & $22.7_{3.5}$\\
    \midrule
    Orig + Fooled (37.7k) & $28.5_{0.9}$ & $45.8_{1.3}$ & $35.0_{0.8}$ & $52.5_{1.0}$ & $20.3_{0.7}$ & $34.9_{1.0}$ & $14.3_{1.0}$ & $19.8_{1.4}$ & $36.7_{1.3}$ & $56.5_{1.5}$ & $15.3_{1.6}$ & $24.3_{2.0}$\\
    Orig + Random (37.7k) & $28.1_{1.5}$ & $45.9_{1.3}$ & $34.1_{1.1}$ & $51.7_{1.1}$ & $19.2_{1.1}$ & $34.1_{1.8}$ & $14.3_{0.8}$ & $20.1_{1.3}$ & $35.6_{1.7}$ & $55.3_{1.4}$ & $15.0_{1.4}$ & $24.5_{2.0}$\\
    Orig + SDC (37.7k) & $\mathbf{30.5_{1.1}}$ & $\mathbf{47.8_{0.8}}$ & $35.8_{1.1}$ & $53.4_{0.8}$ & $\mathbf{23.0_{0.7}}$ & $\mathbf{40.2_{0.7}}$ & $\mathbf{16.5_{0.6}}$ & $\mathbf{22.8_{1.1}}$ & $\mathbf{40.6_{0.6}}$ & $\mathbf{60.7_{0.4}}$ & $\mathbf{18.8_{0.8}}$ & $\mathbf{30.0_{0.8}}$\\
\bottomrule
  \end{tabular}
  \caption{EM and F1 scores of various models evaluated on MRQA dev and test sets. Adversarial results in bold are statistically significant compared to SDC setting and vice versa with $p<0.05.$
  \label{tab:ood_electra_data}}
\end{table*}
\endgroup

\begin{table*}[t!]
\centering
\small
  \begin{tabularx}{16cm}{lX}
    \toprule
    Resource & Examples \\
    \midrule
     & \emph{Lothal [SEP] Lothal ( ) is \textcolor{red}{one of the southernmost cities of the ancient Indus Valley Civilization , located in the Bhāl region ( Ahammedabad District , Dholka Taluk)of the modern state of Gujarāt} and first inhabited 3700 BCE . The meaning of the word Lothal is `` the mount of the dead '' exactly same as that of Mohenjodaro another famous site of Indus Valley civilization . Discovered in 1954 , Lothal was excavated from 13 February 1955 to 19 May 1960 by the Archaeological Survey of India ( ASI ) , the official Indian government agency for the preservation of ancient monuments . According to the ASI , Lothal had the world 's earliest} \newline
    \textbf{What is Lothal and its ancient location?}
    \\
    BERT\textsubscript{fooled} & \emph{One Way or Another [SEP] `` One Way or Another '' is a song by American new wave band Blondie from the album `` Parallel Lines '' . The song was released as the fourth single in the US and Canada as the follow - up to the no . 1 hit `` Heart of Glass '' . " One Way or Another " reached No . 24 on the `` Billboard '' Hot 100 and No . 7 on \textcolor{red}{the `` RPM '' 100 Singles} . Written by Debbie Harry and Nigel Harrison for the band 's third studio album , `` Parallel Lines '' ( 1978 ) , the song was inspired by one of Harry 's ex - boyfriends who stalked her after their breakup . The song was} \newline
    \textbf{Not only did One Way or Another chart on Billboard Hot 100 but it also climbed what other chart?}
 \\
    & \emph{India International Exchange [SEP] The India International Exchange ( INX ) is India 's first international stock exchange , opened in 2017 . It is located at the International Financial Services Centre ( IFSC ) , GIFT City in Gujarat . It is a wholly owned subsidiary of the Bombay Stock Exchange ( BSE ) . The INX will be initially headed by V. Balasubramanian with other staff from \textcolor{red}{the BSE} . It was inaugurated on 9 January 2017 by Indian prime minister Narendra Modi , the trading operations were scheduled to begin on 16 January 2017 . It was claimed to be the world ’s most advanced technological platform with a turn - around time of 4 micro} \newline
    \textbf{Where will the workers of the INX come from?}\\
    \midrule
     & \emph{True Detective ( season 2 ) [SEP] The second season of `` True Detective '' , an American anthology crime drama television series created by \textcolor{red}{Nic Pizzolatto} , began airing on June 21 , 2015 , on the premium cable network HBO . With a principal cast of Colin Farrell , Rachel McAdams , Taylor Kitsch , Kelly Reilly , and Vince Vaughn , the season comprises eight episodes and concluded its initial airing on August 9 , 2015 . The season 's story takes place in California and follows the interweaving stories of officers from three cooperating police departments ; when California Highway Patrol officer and war veteran Paul Woodrugh ( Kitsch )} \newline
    \textbf{Who created True Detective?}
    \\
    BERT\textsubscript{random} & \emph{History of time in the United States [SEP] The history of standard time in the United States began November 18 , 1883 , when United States and Canadian railroads instituted standard time in time zones . Before then , time of day was a local matter , and most cities and towns used some form of \textcolor{red}{local solar time} , maintained by some well - known clock ( for example , on a church steeple or in a jeweler 's window ) . The new standard time system was not immediately embraced by all . Use of standard time gradually increased because of its obvious practical advantages for communication and travel . Standard time in time} \newline
    \textbf{What form of time did most cities and towns use before standard?}
 \\
    & \emph{One Call Away ( Charlie Puth song ) [SEP] `` One Call Away '' is a song by American singer Charlie Puth for his debut album \textcolor{red}{`` Nine Track Mind ''} . It was released on August 20 , 2015 by Atlantic Records as the second single from the album , after the lead single `` Marvin Gaye '' . `` One Call Away '' is a gospel - infused pop soul song . It reached number 12 on the `` Billboard '' Hot 100 , making it Puth 's third top 40 single in the US and his third highest - charting single as a lead artist to date , behind `` We Do n't Talk Anymore '' and} \newline
    \textbf{What is Charlie Puth's first album?}\\
    \midrule
     & \emph{Cap of invisibility [SEP] In classical mythology , \textcolor{red}{the Cap of Invisibility} ( `` ( H)aïdos kuneēn '' in Greek , lit . dog - skin of Hades ) is a helmet or cap that can turn the wearer invisible . It is also known as the Cap of Hades , Helm of Hades , or Helm of Darkness . Wearers of the cap in Greek myths include Athena , the goddess of wisdom , the messenger god Hermes , and the hero Perseus . The Cap of Invisibility enables the user to become invisible to other supernatural entities , functioning much like the cloud of mist that the gods surround themselves in to become undetectable . One ancient} \newline
    \textbf{What is the name given to a cap or helmet that renders the wearer unable to be seen in classical mythology?}
    \\
    SDC & \emph{The Dark Side of the Moon [SEP] The Dark Side of the Moon is the eighth studio album by English rock band Pink Floyd , released on 1 March 1973 by \textcolor{red}{Harvest Records} . It built on ideas explored in Pink Floyd 's earlier recordings and performances , but without the extended instrumentals that characterised their earlier work . A concept album , its themes explore conflict , greed , time , and mental illness , the latter partly inspired by the deteriorating health of founding member Syd Barrett , who left in 1968 . Developed during live performances , Pink Floyd premiered an early version of `` The Dark Side of the Moon} \newline
    \textbf{Which company released the album ``The Dark Side of the Moon''?}
 \\
    & \emph{The Boy in the Striped Pyjamas [SEP] The Boy in the Striped Pyjamas is a 2006 Holocaust novel by Irish novelist John Boyne . Unlike the months of planning Boyne devoted to his other books , he said that he wrote the entire first draft of `` The Boy in the Striped Pyjamas '' in \textcolor{red}{two and a half days} , barely sleeping until he got to the end . He did , however , commit to nearly 20 years of research , reading and researching about the Holocaust as a teenager before the idea for the novel even came to him . As of March 2010 , the novel had sold} \newline
    \textbf{How many days did it take John Boyne to write the first draft of The Boy in the Striped Pyjamas?}
    \\
\bottomrule
  \end{tabularx}
  \caption{Validation set examples of questions in different resources. Correct answers are highlighted in \textcolor{red}{red}.
  \label{tab:examples_bert}\vspace{-2mm}}
\end{table*}

\begin{table*}[t!]
\centering
\small
  \begin{tabularx}{16cm}{lX}
    \toprule
    Resource & Examples \\
    \midrule
     & \emph{Six ( TV series ) [SEP] Six ( stylized as SIX ) is an American television drama series . The series was ordered by the History channel with an eight - episode initial order . The first two episodes were directed by \textcolor{red}{Lesli Linka Glatter} . `` Six '' premiered on January 18 , 2017 . `` Six '' was renewed for a second season of 10 episodes on February 23 , 2017 , which premiered on May 28 , 2018 , with the second new episode airing during its regular timeslot on May 30 , 2018 . On June 29 , History announced they had cancelled the series after two seasons . The series chronicles the operations and daily lives of operators} \newline
    \textbf{Who directed the first two episodes of six?}
    \\
    ELECTRA\textsubscript{fooled} & \emph{Outer space [SEP] Outer space , or just space , is the expanse that exists beyond the Earth and between celestial bodies . Outer space is not completely empty — it is a hard vacuum containing a low density of particles , predominantly a plasma of hydrogen and helium as well as electromagnetic radiation , magnetic fields , neutrinos , dust , and cosmic rays . The baseline temperature , as set by the background radiation from the Big Bang , is . The \textcolor{red}{plasma between galaxies} accounts for about half of the baryonic ( ordinary ) matter in the universe ; it has a number density of less than one hydrogen atom per cubic} \newline
    \textbf{Half of the ordinary matter in the universe is comprised of what?}
 \\
    & \emph{\textcolor{red}{Ode to Billie Joe} [SEP] `` Ode to Billie Joe '' is a song written and recorded by Bobbie Gentry , a singer - songwriter from Chickasaw County , Mississippi . The single , released on July 10 , 1967 , was a number - one hit in the US and a big international seller . `` Billboard '' ranked the record as the No . 3 song of the year . It generated eight Grammy nominations , resulting in three wins for Gentry and one for arranger Jimmie Haskell . `` Ode to Billie Joe '' has since made `` Rolling Stone'' 's lists of the `` 500 Greatest Songs of All Time '' and the `` 100 Greatest Country Songs of All Time '' and `` Pitchfork''} \newline
    \textbf{What did ``Billboard'' rank as the No. 3 song of the year in 1967?}\\
    \midrule
     & \emph{Sagrada Família [SEP] The (; ; ) is a large unfinished \textcolor{red}{Roman Catholic church} in Barcelona , designed by Catalan architect Antoni Gaudí ( 1852–1926 ) . Gaudí  's work on the building is part of a UNESCO World Heritage Site , and in November 2010 Pope Benedict XVI consecrated and proclaimed it a minor basilica , as distinct from a cathedral , which must be the seat of a bishop . In 1882 , construction of Sagrada Família started under architect Francisco de Paula del Villar . In 1883 , when Villar resigned , Gaudí took over as chief architect , transforming the project with his architectural and engineering style} \newline
    \textbf{What kind of unfinished church is the Sagrada Família?}
    \\
    ELECTRA\textsubscript{random} & \emph{Loyola Ramblers men 's basketball [SEP] The Loyola Ramblers men 's basketball team represents Loyola University Chicago in Chicago , Illinois . The Ramblers joined the Missouri Valley Conference on \textcolor{red}{July 1 , 2013} , ending a 34-season tenure as charter members of the Horizon League . In 1963 , Loyola won the 1963 NCAA Men 's Division I Basketball Tournament ( then the `` NCAA University Division '' ) men 's basketball national championship under the leadership of All - American Jerry Harkness , defeating two - time defending champion Cincinnati 60–58 in overtime in the title game . All five starters for the Ramblers played the entire championship game without substitution . Surviving team members were} \newline
    \textbf{When did the Ramblers join the Missouri Valley Conference?}
 \\
    & \emph{The Walking Dead ( season 7 ) [SEP] The seventh season of `` The Walking Dead '' , an American post - apocalyptic horror television series on AMC , premiered on October 23 , 2016 , and concluded on April 2 , 2017 , consisting of 16 episodes . Developed for television by Frank Darabont , the series is based on the eponymous series of \textcolor{red}{comic books} by Robert Kirkman , Tony Moore , and Charlie Adlard . The executive producers are Kirkman , David Alpert , Scott M. Gimple , Greg Nicotero , Tom Luse , and Gale Anne Hurd , with Gimple as showrunner for the fourth consecutive season . The seventh season received} \newline
    \textbf{What was the Walking Dead's original source material?}\\
    \midrule
     & \emph{Southern California Edison [SEP] Southern California Edison ( or SCE Corp ) , the largest subsidiary of Edison International , is the primary electricity supply company for much of Southern California . It provides \textcolor{red}{14 million} people with electricity across a service territory of approximately 50,000 square miles . However , the Los Angeles Department of Water and Power , San Diego Gas \& Electric , Imperial Irrigation District , and some smaller municipal utilities serve substantial portions of the southern California territory . The northern part of the state is generally served by the Pacific Gas \& Electric} \newline
    \textbf{How many people does SCE Corp provide with electricity?}
    \\
    SDC & \emph{Do n't Go Away [SEP] `` Do n't Go Away '' is a song by the English rock band Oasis from their third album , `` \textcolor{red}{Be Here Now} '' , written by the band 's lead guitarist Noel Gallagher . The song was released as a commercial single only in Japan , peaking at number 48 on the Oricon chart , and as a promotional single in the United States , Japan and Europe . In the United States it was a success , hitting \# 5 on the `` Billboard '' Hot Modern Rock Tracks chart in late 1997 . It was the band 's last major hit in the United} \newline
    \textbf{What Oasis album is ``Don't go away'' from?}
 \\
    & \emph{India national cricket team [SEP] The India national cricket team , also known as Team India and Men in Blue , is governed by the Board of Control for Cricket in India ( BCCI ) , and is a full member of the International Cricket Council ( ICC ) with Test , \textcolor{red}{One Day International} ( ODI ) and Twenty20 International ( T20I ) status . Although cricket was introduced to India by European merchant sailors in the 18th century , and the first cricket club was established in Calcutta ( currently known as Kolkata ) in 1792 , India 's national cricket team did not play its first Test match until 25 June 1932 at Lord 's } \newline
    \textbf{What does ODI stand for?}\\
\bottomrule
  \end{tabularx}
  \caption{Validation set examples of questions in different resources. Correct answers are highlighted in \textcolor{red}{red}.
  \label{tab:examples_electra}\vspace{-2mm}}
\end{table*}
\end{document}